\documentclass[11pt]{article}

\usepackage[T1]{fontenc}
\usepackage{lmodern}
\usepackage[margin=1in]{geometry}
\usepackage{microtype}
\usepackage{amsmath,amssymb}
\usepackage{graphicx}
\usepackage{booktabs}
\usepackage{longtable}
\usepackage{tabularx}
\usepackage{array}
\usepackage{enumitem}
\usepackage{float}
\usepackage{xcolor}
\usepackage{tikz}
\usetikzlibrary{arrows.meta,backgrounds,calc,fit,positioning,shapes.geometric}
\usepackage[authoryear,square,sort]{natbib}
\usepackage{xurl}
\usepackage[hidelinks]{hyperref}
\hypersetup{
  pdftitle={From Matching Models to Recruiting Agents: A Systematized Narrative Review of AI Recruitment Systems, Evaluation, and Governance},
  pdfauthor={Ziyi Zhao and Guanzheng Wei},
  pdfkeywords={recruitment AI, job recommendation, talent search, large language models, agents, evaluation, governance}
}

\definecolor{stageblue}{HTML}{2457A6}
\definecolor{stagegreen}{HTML}{2D7A46}
\definecolor{stageorange}{HTML}{A85B16}
\definecolor{stagepurple}{HTML}{6B4C9A}
\definecolor{stagecyan}{HTML}{197D8B}
\definecolor{stagered}{HTML}{A43A3A}
\definecolor{lightgray}{HTML}{F3F4F6}
\definecolor{palegreen}{HTML}{EAF5ED}
\definecolor{paleblue}{HTML}{EAF1FB}
\definecolor{paleorange}{HTML}{FBEFE4}
\definecolor{palepurple}{HTML}{F1ECF8}
\definecolor{palecyan}{HTML}{E8F5F6}
\definecolor{palered}{HTML}{F9EAEA}

\newcolumntype{Y}{>{\raggedright\arraybackslash}X}
\newcolumntype{P}[1]{>{\raggedright\arraybackslash}p{#1}}

\newcommand{\reviewcutoff}{23 July 2026}
\newcommand{\supplementcutoff}{28 July 2026}
\newcommand{\forwardcutoff}{2 September 2026}
\newcommand{\legalcutoff}{2 September 2026}

\title{From Matching Models to Recruiting Agents:\\
A Systematized Narrative Review of AI Recruitment Systems, Evaluation, and Governance}
\author{%
\textbf{Ziyi Zhao}$^{1,\dagger}$ and \textbf{Guanzheng Wei}$^{2,\dagger}$\\[0.45em]
\small $^{1}$University of Chinese Academy of Social Sciences \quad
\href{mailto:s20241040534@ucass.edu.cn}{\texttt{s20241040534@ucass.edu.cn}}\\[-0.05em]
\small $^{2}$Southwest University \quad
\href{mailto:wgz021228@email.swu.edu.cn}{\texttt{wgz021228@email.swu.edu.cn}}\\[0.25em]
\small $^{\dagger}$These authors contributed equally.}
\date{}

\setlist[itemize]{leftmargin=*,topsep=2pt,itemsep=1pt}
\setlist[enumerate]{leftmargin=*,topsep=2pt,itemsep=1pt}

\begin{document}
\maketitle

\begin{abstract}
Artificial intelligence in recruitment has shifted the object being automated
from profile pairs and ranked lists to multi-stage workflows that retrieve
evidence, compare candidates, and support or execute actions. This systematized
narrative review traces that development from bilateral retrieval and
behavioral ranking through neural person--job matching, large-language-model
(LLM) components, and tool-using recruiting agents. Using a purposive search
and coding protocol updated through 23 July 2026, with targeted
cross-disciplinary, regional, and forward updates through 2 September 2026,
we organize 40 representative works alongside supporting industrial,
regulatory, and statutory sources. The map is a structured synthesis, not a
prevalence estimate. We
analyze three coupled transitions: from similarity to reciprocal suitability,
from a model to a compound workflow, and from offline prediction to evidence-
and productivity-aligned evaluation. Across document understanding, retrieval,
ranking, assessment, interviewing, sourcing, and human handoff, we distinguish
field-, pair-, list-, case-, trajectory-, and outcome-level evidence.
Persistent gaps arise because behavioral labels confound exposure, preference,
and qualification; private and synthetic data limit external validity;
final-output scores conceal pipeline failures; and, within the coded set,
privacy is not directly evaluated and no row jointly evaluates utility,
fairness, privacy, and security. These observations describe the coded set
rather than the field as a whole. We
therefore introduce a staged mapping from evaluation evidence to the
strongest defensible claim, together with an agenda for reciprocal,
evidence-grounded, temporally controlled, selective, and auditable systems.
Progress should be judged by
whether the complete workflow retrieves the right evidence, preserves
uncertainty, supports contestable decisions, and improves outcomes under
explicit cost and risk constraints.
\end{abstract}

\section{Introduction}
\label{sec:introduction}

Recruitment is often described as a matching problem, but the phrase hides
several distinct decisions. A platform may retrieve jobs for a seeker, retrieve
candidates for a recruiter, estimate whether a person satisfies stated
requirements, predict whether either side will respond, recommend an interview,
or automate the work needed to produce a defensible shortlist. These decisions
have different inputs, losses, users, and consequences. Treating all of them as
one ``AI match score'' makes systems difficult to compare and easier to
overclaim.

Throughout this review, \emph{matching} is an umbrella term for connecting
people and jobs; \emph{qualification} denotes employer-side evidence against
job-related criteria; \emph{preference} denotes candidate- or employer-side
willingness; and \emph{selection} denotes filtering or assessment that
materially affects access to employment. We reserve \emph{recruiting agent} for
a system with case state, tool-mediated observation, and a policy that adapts
its next step to an environment during execution. We use \emph{action-capable
recruiting agent} only when that policy can also mutate an external system or
initiate consequential communication under explicit authority.

The modern search-and-recommender lineage of recruitment AI adapted information
retrieval and recommender systems to a
two-sided market. An early bilateral formulation explicitly modeled both
candidate and recruiter preferences~\citep{malinowski-etal-2006-bilateral};
later work learned job transitions from public profiles
\citep{paparrizos-etal-2011-machine} and deployed multi-tier recommendation
cascades at platform scale~\citep{kenthapadi-etal-2017-linkedin}. Deep models
then learned joint representations from resumes, job descriptions, and
historical applications~\citep{zhu-etal-2018-pjfnn,qin-etal-2018-apjfnn}, while
co-teaching, graphs, and contrastive objectives addressed noisy and sparse
interaction data~\citep{bian-etal-2020-coteaching,yu-etal-2024-confit}.

LLMs expanded the design space again. They can rewrite or synthesize training
examples, extract latent requirements, generate an idealized job or resume,
re-rank a shortlist, explain criterion-level evidence, and operate tools over
the open web~\citep{zheng-etal-2026-girl,du-etal-2024-job-gan,
yu-etal-2025-confit,yu-etal-2026-confit-v3}. Multi-module assessment
prototypes, sometimes self-described as agentic, combine resumes, job
descriptions, interview transcripts, and recruiter feedback into rubric-based
workflows~\citep{gan-etal-2024-application,
yuksel-etal-2026-agentic}. In xbench, recruitment is no longer represented by a
single text pair: agents execute headhunting tasks involving company mapping,
information retrieval, and talent sourcing in a dynamic environment
\citep{chen-etal-2025-xbench}.

This change creates a measurement mismatch. Binary classification metrics can
be useful for a fixed candidate--job pair but say little about corpus-level
reachability. Ranking metrics measure order but not whether the underlying
evidence is true. Click and application rates capture marketplace behavior but
conflate exposure, preference, qualification, and interface effects. End-to-end
agent success may hide unsafe intermediate actions, stale facts, or incomplete
coverage. Conversely, a system can help a recruiter without autonomously
making a valid employment decision. The unit of evaluation must therefore
expand with the unit of automation.

\begin{figure}[H]
\centering
\begingroup
\hyphenpenalty=10000\relax
\exhyphenpenalty=10000\relax
\resizebox{0.98\textwidth}{!}{%
\begin{tikzpicture}[x=1cm,y=1cm,
  metric/.style={rounded corners=2pt, draw=#1!72, fill=#1!7, line width=0.8pt,
    minimum width=4.35cm, minimum height=0.74cm, text width=4.02cm,
    align=left, inner sep=4pt, font=\fontsize{7.4}{8.1}\selectfont},
  unit/.style={rounded corners=8pt, draw=#1, fill=#1, text=white,
    line width=1.0pt, minimum width=2.45cm, minimum height=0.74cm,
    text width=2.20cm, align=center, inner sep=3pt,
    font=\fontsize{7.8}{8.4}\selectfont\bfseries},
  claim/.style={rounded corners=2pt, draw=black!28, fill=lightgray,
    minimum width=6.10cm, minimum height=0.74cm, text width=5.72cm,
    align=left, inner sep=4pt, font=\fontsize{7.4}{8.1}\selectfont},
  head/.style={font=\fontsize{9.0}{9.6}\selectfont\bfseries,
    text=black!68, align=center},
  connector/.style={-{Latex[length=1.8mm,width=1.3mm]}, line width=0.9pt,
    draw=black!46, shorten >=0.6pt, shorten <=0.6pt}
]

\node[rounded corners=3pt, draw=black!22, fill=lightgray,
  font=\fontsize{7.4}{8.1}\selectfont\bfseries, text=black!62,
  inner xsep=8pt, inner ysep=4pt]
  at (6.95,4.75)
  {$\uparrow$ Broader system scope requires new evidence; causal claims require identified designs};

\node[head] at (2.03,3.90) {Observable metrics};
\node[head] at (5.95,3.90) {Evaluation unit};
\node[head] at (10.745,3.90) {Maximum supported claim};

\node[metric=stagered] (m6) at (2.03,3.15)
  {funnel, quality, retention,\\adverse impact, causal design};
\node[unit=stagered] (u6) at (5.95,3.15) {Outcome};
\node[claim] (c6) at (10.745,3.15)
  {Deployment outcomes are observed; causal\\change requires an identified design.};

\node[metric=stagecyan] (m5) at (2.03,1.97)
  {completion, safe actions, recovery,\\latency, cost};
\node[unit=stagecyan] (u5) at (5.95,1.97) {Trajectory};
\node[claim] (c5) at (10.745,1.97)
  {The workflow runs reliably in the\\tested environment.};

\node[metric=stagepurple] (m4) at (2.03,0.79)
  {evidence precision / recall,\\adjudication, risk--coverage};
\node[unit=stagepurple] (u4) at (5.95,0.79) {Case};
\node[claim] (c4) at (10.745,0.79)
  {A case report is grounded, selective,\\and reviewable.};

\node[metric=stageorange] (m3) at (2.03,-0.39)
  {Recall@$K$, nDCG@$K$, coverage,\\exposure, stability};
\node[unit=stageorange] (u3) at (5.95,-0.39) {List};
\node[claim] (c3) at (10.745,-0.39)
  {Relevant items are reachable and\\appropriately ordered.};

\node[metric=stagegreen] (m2) at (2.03,-1.57)
  {AUC, calibration, criterion accuracy,\\abstention};
\node[unit=stagegreen] (u2) at (5.95,-1.57) {Pair};
\node[claim] (c2) at (10.745,-1.57)
  {A defined candidate--job label can be\\predicted.};

\node[metric=stageblue] (m1) at (2.03,-2.75)
  {EM, F1, schema errors, span,\\and source provenance};
\node[unit=stageblue] (u1) at (5.95,-2.75) {Field / document};
\node[claim] (c1) at (10.745,-2.75)
  {Extraction is correct on the sampled\\documents.};

\foreach \m/\u/\c in {m6/u6/c6,m5/u5/c5,m4/u4/c4,m3/u3/c3,m2/u2/c2,m1/u1/c1} {
  \draw[connector] (\m.east) -- (\u.west);
  \draw[connector] (\u.east) -- (\c.west);
}
\foreach \lower/\upper in {u1/u2,u2/u3,u3/u4,u4/u5,u5/u6}
  {\draw[connector] (\lower.north) -- (\upper.south);}

\node[rounded corners=3pt, draw=stagered!72, fill=palered,
  font=\fontsize{7.4}{8.1}\selectfont\bfseries, text=stagered,
  inner sep=4pt]
  at (6.95,-3.65)
  {No automatic upward propagation: evidence must be re-established at every unit};

\end{tikzpicture}%
}
\endgroup
\caption{Evaluation units and claim ceilings. The staircase connects observable
metrics to the strongest claim each unit can support; moving upward requires
new evidence rather than a more persuasive interpretation of a lower-level
score. The order represents broader system scope and consequence, not a
universal ranking of study quality. Schema-valid extraction does not imply list
recall, and successful task
execution does not establish employment-selection validity. Grounded generation
and selective prediction remain useful component methods
~\citep{gao-etal-2023-enabling,ru-etal-2024-ragchecker,
buchmann-etal-2024-attribute,muhamed-etal-2026-refusalbench}.}
\label{fig:evaluation-ladder}
\end{figure}
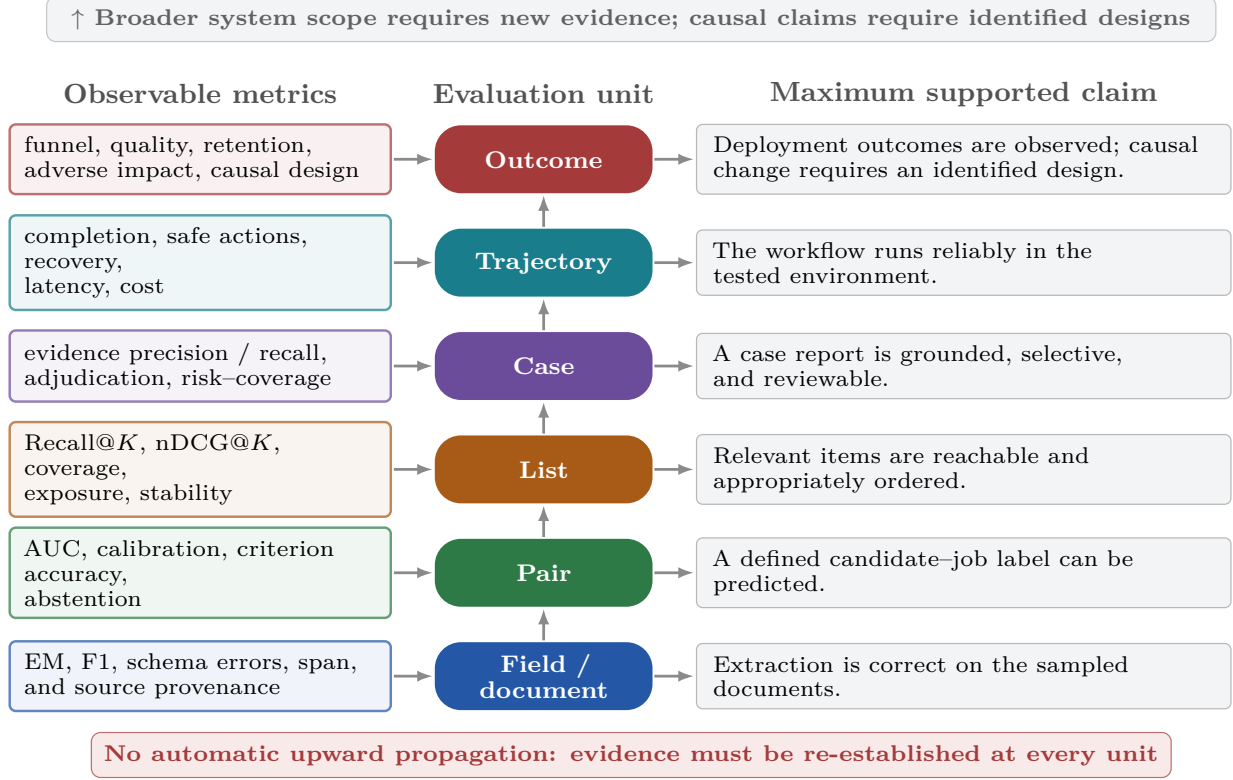

This unit-to-evidence alignment is the paper's core insight: as automation
expands from isolated pairs to multi-stage workflows, evaluation must expand
from pair prediction to list coverage, case grounding, trajectory safety, and
outcome evidence. Figure~\ref{fig:evaluation-ladder} states this claim and the
maximum defensible conclusion at each evaluation unit.

\paragraph{Contributions.}
This systematized narrative review makes two core contributions. First, it
connects recruitment stage, automation scope, and evaluation unit in one
framework that separates extraction, recall, filtering, ranking, judgment,
action, and outcome. Second, it develops a staged evidence-to-claim mapping
that distinguishes predictive performance, construct validity, business
utility, factual grounding, allocational risk, and outcome evidence.

Three supporting artifacts make those contributions inspectable: (i) a
reconstruction of five overlapping layers in the modern search-and-recommender
lineage; (ii) a triangulated evidence map spanning academic research,
industrial papers, corporate disclosures, regulator audits, and selected
regional rules; and (iii) a failure-to-intervention map and research agenda
that separate recruitment-tested interventions, transferable methods, and
unvalidated designs.

Existing reviews illuminate complementary slices of this landscape.
Table~\ref{tab:related-surveys} makes the distinction explicit. Job-recommender
surveys organize retrieval and recommendation challenges; skill-extraction and
HR-NLP surveys cover important components; fairness reviews examine
allocational risk; and an LLM scoping review maps recent hiring-decision uses.
Our emphasis is the evidence chain connecting the pre-LLM search and
marketplace lineage to LLM components, recruiting agents, and
profession-aligned evaluation. Table~\ref{tab:related-surveys} compares the
organizing questions rather than claiming broader or better coverage: the
extension developed here is one cross-stage mapping from automated artifact to
evaluation unit, authority, and claim ceiling.

\begin{table}[t]
\centering
\small
\setlength{\tabcolsep}{4pt}
\caption{Positioning relative to representative adjacent reviews.}
\label{tab:related-surveys}
\begin{tabularx}{\textwidth}{@{}P{4.0cm}P{5.0cm}Y@{}}
\toprule
\textbf{Review} & \textbf{Primary organizing emphasis} &
\textbf{Extension developed here} \\
\midrule
Job recommender systems:\newline
\citeauthor{de-ruijt-bhulai-2021-review}
(\citeyear{de-ruijt-bhulai-2021-review}, preprint);\newline
\citeauthor{mashayekhi-etal-2024-challenge}
(\citeyear{mashayekhi-etal-2024-challenge}) & Recommendation architectures, data, and
evaluation challenges & Connect ranking to upstream documents, downstream
case work, and agent trajectories \\
Computational skill extraction~\citep{senger-etal-2024-skill-survey} &
Extraction and classification from job postings & Place document intelligence
inside a full retrieval--decision pipeline \\
NLP for human resources~\citep{otani-etal-2025-natural} & NLP tasks across HR
workflows & Trace changes in automation object, authority, and claim ceiling \\
Algorithmic-hiring fairness~\citep{fabris-etal-2025-fairness-survey} &
Multidisciplinary bias and fairness evidence & Audit disparity jointly with
stage utility, grounding, security, and human correction \\
Disability and AI-mediated hiring~\citep{guler-zhu-2026-disability} &
Accessibility opportunities, disclosure tensions, and ableist risk & Connect
accommodation and disability-centered design to each evaluation stage \\
LLMs in hiring decisions~\citep{tripathi-etal-2026-mapping} & Cross-disciplinary
map of LLM use in hiring & Link the pre-LLM lineage to compound and tool-using
systems with trajectory and lifecycle evaluation \\
\bottomrule
\end{tabularx}
\end{table}

\begin{figure}[H]
\centering
\begingroup
\hyphenpenalty=10000\relax
\exhyphenpenalty=10000\relax
\renewcommand{\tiny}{\fontsize{7.0}{7.6}\selectfont}
\newcommand{\milestone}[6]{%
  {\textcolor{#1}{\textbf{#2}}\enspace\textbf{#3}\\[-0.4pt]
   \textbf{#4}\\[0.8pt]
   {\tiny #5\\[-0.2pt]#6}}}
\resizebox{\textwidth}{!}{%
\begin{tikzpicture}[x=1cm,y=1cm,
  era/.style={rounded corners=3pt, draw=#1, fill=#1!10, line width=1pt,
    minimum width=2.85cm, minimum height=1.10cm, text width=2.58cm,
    align=center, inner sep=3pt, font=\scriptsize},
  card/.style={rounded corners=2pt, draw=#1!75, fill=white, line width=0.75pt,
    minimum width=2.85cm, minimum height=1.84cm, text width=2.55cm,
    align=left, inner sep=4pt, font=\scriptsize},
  year/.style={rounded corners=6pt, draw=stageblue!75, fill=white,
    line width=0.9pt, minimum width=1.55cm, minimum height=0.42cm,
    font=\scriptsize\bfseries, text=stageblue},
  link/.style={line width=0.65pt, draw=black!40}
]

\node[era=stageblue]   at (1.40,3.45) {\textbf{Rules and bilateral IR}\\2006--2012};
\node[era=stagegreen]  at (4.60,3.45) {\textbf{Behavioral ranking}\\2013--2019};
\node[era=stageorange] at (7.80,3.45) {\textbf{Neural matching}\\2018--2023};
\node[era=stagepurple] at (11.00,3.45) {\textbf{LLM pipelines}\\2023--2025};
\node[era=stagecyan]   at (14.20,3.45) {\textbf{Recruiting agents}\\2025--2026};

\node[card=stageblue] (m11) at (1.40,1.82)
  {\milestone{stageblue}{P}{Bilateral}{recommendation}
    {two-sided preference}{Malinowski et al.}};
\node[card=stageblue] (m12) at (1.40,-1.28)
  {\milestone{stageblue}{T}{Career}{transitions}
    {observed next job}{Paparrizos et al.}};

\node[card=stagegreen] (m21) at (4.60,1.82)
  {\milestone{stagegreen}{R}{Industrial}{retrieval}
    {multi-tier ranker}{LinkedIn (2017)}};
\node[card=stagegreen] (m22) at (4.60,-1.28)
  {\milestone{stagegreen}{F}{Talent search}{and fairness}
    {entity-aware serving}{fair re-ranking}};

\node[card=stageorange] (m31) at (7.80,1.82)
  {\milestone{stageorange}{M}{Neural}{person--job fit}
    {semantic alignment}{PJFNN (2018)}};
\node[card=stageorange] (m32) at (7.80,-1.28)
  {\milestone{stageorange}{M}{Reciprocal}{graph matching}
    {two-way interaction}{DPGNN (2022)}};

\node[card=stagepurple] (m41) at (11.00,1.82)
  {\milestone{stagepurple}{L}{LLM retrieval}{and cases}
    {dense retrieval}{ConFit, 2024--25}};
\node[card=stagepurple] (m42) at (11.00,-1.28)
  {\milestone{stagepurple}{W}{LLM}{workflow}
    {modular screening}{Gan et al. (2024)}};

\node[card=stagecyan] (m51) at (14.20,1.82)
  {\milestone{stagecyan}{A}{Adaptive job}{recommendation}
    {AdaptJobRec}{(2026)}};
\node[card=stagecyan] (m52) at (14.20,-1.28)
  {\milestone{stagecyan}{B}{Agent}{evaluation}
    {xbench (2025)}{PeopleSearchBench}};

\draw[-{Latex[length=3.0mm]}, line width=2.3pt, draw=stageblue!58,
  rounded corners=7pt]
  (0.05,0.14) -- (15.75,0.14);

\node[year] (y1) at (1.40,0.14) {2006--11};
\node[year] (y2) at (4.60,0.14) {2017--19};
\node[year] (y3) at (7.80,0.14) {2018--22};
\node[year] (y4) at (11.00,0.14) {2024--25};
\node[year] (y5) at (14.20,0.14) {2025--26};
\foreach \top/\yr/\bottom in {m11/y1/m12,m21/y2/m22,m31/y3/m32,m41/y4/m42,m51/y5/m52} {
  \draw[link] (\top.south) -- (\yr.north);
  \draw[link] (\yr.south) -- (\bottom.north);
}

\node[draw=black!22, rounded corners=3pt, fill=lightgray,
  minimum width=15.65cm, minimum height=0.94cm, align=center,
  font=\fontsize{6.8}{7.4}\selectfont] at (7.80,-2.88)
  {\textcolor{stageblue}{\textbf{P/T}: preference and transition}\quad
   \textcolor{stagegreen}{\textbf{R/F}: retrieval, ranking, fairness}\quad
   \textcolor{stageorange}{\textbf{M}: representation and matching}\\[1pt]
   \textcolor{stagepurple}{\textbf{L/W}: LLM retrieval and workflow}\quad
   \textcolor{stagecyan}{\textbf{A/B}: adaptive recommendation and agent evaluation}};

\end{tikzpicture}%
}
\endgroup
\caption{Timeline of representative recruitment-AI systems and evidence
transitions. The eras overlap: production systems retain lexical recall, hard
filters, and human review while adding neural models, LLM evidence modules, and
tool-using agents. Milestones include bilateral recommendation
~\citep{malinowski-etal-2006-bilateral}, industrial ranking
~\citep{kenthapadi-etal-2017-linkedin}, neural person--job fit
~\citep{zhu-etal-2018-pjfnn}, LLM retrieval
~\citep{yu-etal-2024-confit,yu-etal-2025-confit}, modular screening
~\citep{gan-etal-2024-application}, adaptive recommendation
~\citep{wang-etal-2026-adaptjobrec}, and professional-agent evaluation
~\citep{chen-etal-2025-xbench,shi-etal-2026-peoplesearchbench}.}
\label{fig:evolution}
\end{figure}
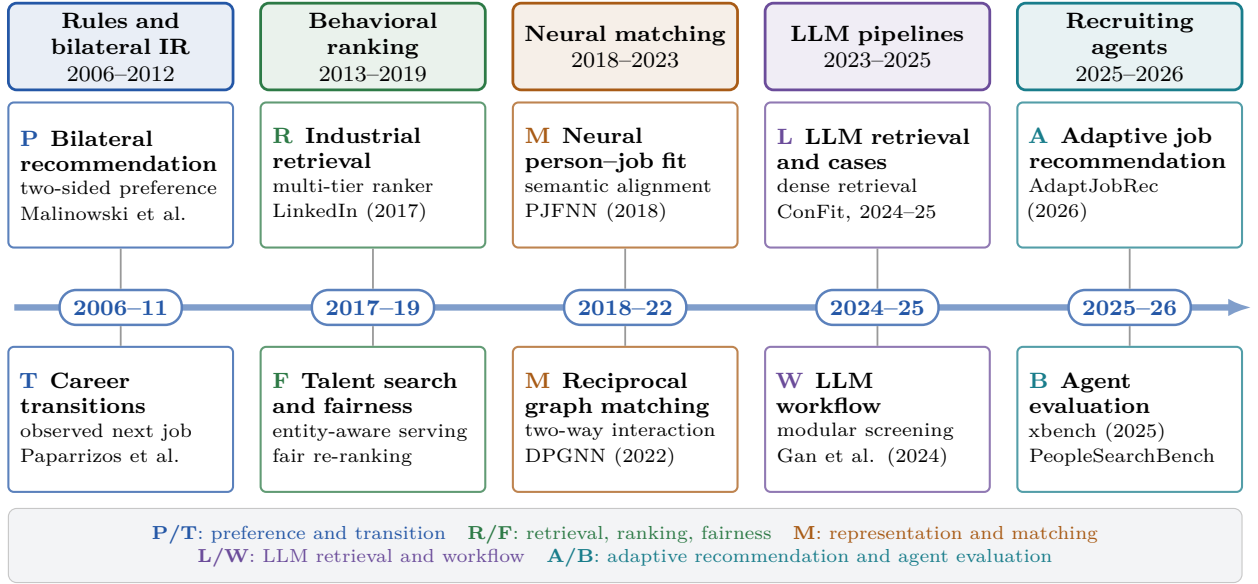

Figure~\ref{fig:evolution} previews the organizing argument. The dominant
artifact expands from a profile pair, to a ranked list, to a case trajectory;
the corresponding evidence must expand from similarity or behavioral labels to
coverage, grounding, action safety, and downstream outcomes.

The scope is recruitment and selection up to the handoff into employment. We
include job recommendation, candidate search, screening, interview support,
and recruiting communication. Post-hire HR analytics are included only when
they establish methods or evidence directly relevant to hiring. The regional
comparison is deliberately limited to selected jurisdictions with verifiable
primary sources; it is not a global legal census.

\section{Review Protocol and Evidence Boundaries}
\label{sec:protocol}

This manuscript is a \emph{systematized narrative review}. It uses an explicit
search and coding protocol, but it does not claim exhaustive coverage of every
database or satisfy all requirements of a registered Preferred Reporting Items
for Systematic Reviews and Meta-Analyses (PRISMA) systematic review.
That distinction matters in a fast-moving area where conference proceedings,
preprints, product systems, standards, and regulation evolve on different
timelines.

\subsection{Search strategy}

The primary technical search was updated through \reviewcutoff. Searches combined recruitment
terms (\emph{recruitment}, \emph{hiring}, \emph{resume}, \emph{CV},
\emph{job recommendation}, \emph{person--job fit}, \emph{talent search},
\emph{interview}) with method terms (\emph{information retrieval},
\emph{learning to rank}, \emph{recommender system}, \emph{transformer},
\emph{large language model}, \emph{agent}, \emph{benchmark}, \emph{fairness},
and \emph{prompt injection}). Primary discovery sources included the ACM
Digital Library, ACL Anthology, AAAI proceedings, IEEE proceedings records,
arXiv, official journal and publisher pages, and official regulator or benchmark
pages. Backward and forward
snowballing began from recent HR-NLP and job-recommendation surveys
\citep{de-ruijt-bhulai-2021-review,mashayekhi-etal-2024-challenge,
otani-etal-2025-natural,tripathi-etal-2026-mapping,
fabris-etal-2025-fairness-survey}.

For reproducibility, the normalized Boolean templates used in the final update
were as follows; field restrictions and punctuation were adapted to each
database while preserving the logical groups.

\begin{enumerate}[label=(\Alph*),leftmargin=2.2em]
  \item (recruit* \textsc{or} hir* \textsc{or} resume \textsc{or} CV
  \textsc{or} ``job recommendation'' \textsc{or} ``person--job fit''
  \textsc{or} ``talent search'' \textsc{or} interview) \textsc{and}
  (``information retrieval'' \textsc{or} ``learning to rank'' \textsc{or}
  recommend* \textsc{or} transformer \textsc{or} ``large language model''
  \textsc{or} agent \textsc{or} benchmark \textsc{or} fairness \textsc{or}
  ``prompt injection'')
  \item (``employee selection'' \textsc{or} ``personnel selection''
  \textsc{or} ``structured interview'' \textsc{or} psychometric*
  \textsc{or} ``job analysis'') \textsc{and} (validity \textsc{or}
  reliability \textsc{or} fairness \textsc{or} ``differential prediction''
  \textsc{or} audit*)
  \item (recruit* \textsc{or} hir* \textsc{or} ``employment decision'')
  \textsc{and} (deploy* \textsc{or} ``online experiment'' \textsc{or}
  regulation \textsc{or} law \textsc{or} audit \textsc{or} ``automated
  decision'')
\end{enumerate}

To reduce the computer-science bias of the primary discovery sources, a
targeted supplementary update was completed on \supplementcutoff{} over official
publisher and journal pages in personnel selection, applicant reactions, and
labor economics. It used combinations of \emph{personnel selection},
\emph{applicant reactions}, \emph{candidate experience}, \emph{organizational
justice}, \emph{job testing}, \emph{manager discretion}, and employment
outcomes such as tenure, productivity, fill rate, and subgroup hiring. A
separate Chinese-language and China-situated query combined terms for AI
recruitment, AI interviews, algorithmic recruitment, and job advertisements
with terms for applicant reactions, fairness perceptions, acceptance,
gendered wording, applications, and callbacks. This was a targeted
cross-disciplinary and non-English supplement, not an exhaustive PsycINFO,
Scopus, Web of Science, CNKI, or Wanfang census. The exact normalized queries,
source surfaces, retained citations, and boundaries are recorded in
\texttt{supplementary\_search\_log.csv}.

To reduce the blind spot between published models and actual deployment, a
second targeted search covered official company engineering pages, product
announcements, annual reports, enacted law, regulator guidance, and regulator
audit reports. These sources were admitted only when an accountable
organization, publication date, and stable first-party page or document could
be identified. Corporate materials are evidence of disclosed product design,
scale, or business priorities; they are not treated as independent validation
of effectiveness. Regulator audits can reveal operational failure modes and
remediation, but may anonymize providers and omit model-level detail.

A targeted forward update then covered 24 July through \forwardcutoff{}. It
queried the arXiv API and Crossref for recruitment, hiring, resume screening,
job recommendation, candidate sourcing, automated interviews, personnel
selection, applicant reactions, fairness, and agent security. Candidate records
were verified on arXiv, publisher, regulator, government, company, or securities-
exchange pages. We retained items that changed the automated artifact, evidence
level, intervention family, regional coverage, or legal-status boundary. The
update was not a retrospective rerun of every database-specific primary query,
so \reviewcutoff{} remains the primary-search cutoff rather than being silently
advanced. Rows S6--S8 of \texttt{supplementary\_search\_log.csv} record the
forward-update surfaces, normalized queries, retained sources, and exclusion
boundaries. Row S9 separately records high-relevance items whose identifiable
primary-record timestamp predates the primary cutoff but that were recovered
only during this audit. A submission timestamp does not imply that a record was
publicly searchable on that date. The legal-status verification date is
\legalcutoff{}.

We included work that (i) directly studies a recruitment workflow or a
methodological dependency such as resume parsing, (ii) provides a verifiable
paper or official primary source, and (iii) supplies enough information to code
the task, data, method, metric, and limitation. Peer-reviewed main-track,
findings, industry, demonstration, workshop, and journal work are not treated
as equivalent evidence. Preprints are retained when they define a recent method
or benchmark unavailable in archival proceedings, but are labeled as such in
the synthesis.

The representative map retains works that change at least one of four things:
the automated system object, the learning signal, the evaluation unit, or the
risk surface. Near-duplicate model variants were not retained merely to
increase paper counts. Earlier work anchors a technical lineage; recent
benchmarks or demonstrations are retained when they introduce a new system
object or evidence question, not as proof that the capability is mature. This
selection rule keeps industrial, validity, fairness, and security evidence
visible when conventional NLP venue searches would miss it
~\citep{raghavan-etal-2020-mitigating}.

This expansion increases source diversity rather than converting the review
into an exhaustive census. Discovery was iterative, and no frozen historical
export was preserved for every database. We therefore do not retrospectively
invent hit counts, deduplication counts, exclusion reasons, dual-screening
statistics, or inter-rater agreement. Search stopped by author judgment when
additional sources repeated an already represented system object, signal,
evaluation unit, solution family, or jurisdictional constraint. This is a
judgmental stopping rule, not evidence of field-wide saturation. The 40 coded
works form a structured trace of the synthesis and its contrasts, not a
prevalence estimate or a statistically representative sample.

The selected evidence also does not support a defensible meta-analysis. Tasks,
candidate pools, labels, outcome definitions, and designs range from offline
retrieval benchmarks and qualitative interviews to field experiments,
regulator audits, and statutory materials; they do not share a common effect
size. A pooled estimate would therefore create false comparability. Quantitative
synthesis would require a narrower, prospectively registered question with
homogeneous interventions and outcomes.

For the same reason, we do not assign one numerical quality score across
academic experiments, industry reports, corporate disclosures, audits, and
law. Instead, source type and claim boundary are coded explicitly. This
source-specific appraisal is not a substitute for a formal risk-of-bias
assessment when a future review asks a narrower causal-effect question.

\paragraph{Use of generative-AI tools.}
Generative-AI tools assisted search-term expansion, source discovery,
provisional coding, language editing, and LaTeX and figure quality assurance.
Their outputs were treated as proposals rather than independent evidence:
claims and bibliographic records were checked against identifiable source
documents. The human authors retain responsibility for eligibility decisions,
interpretation, citations, and the final manuscript.

\subsection{Coding dimensions}

The analytical framework and the item-level codebook serve different purposes.
The framework in Section~\ref{sec:framework} compares systems on recruitment
stage, automation scope, and evaluation unit. In the companion CSV,
\texttt{citation\_key} is the record identifier; controlled tags and
source-specific descriptive fields are intentionally distinguished. The
remaining columns record the following author-coded descriptors:

\begin{enumerate}
  \item \textbf{Publication year}: the year of the cited version.
  \item \textbf{Automated artifact}: one or more tags from
  \texttt{field\_document}, \texttt{pair}, \texttt{list}, \texttt{case}, and
  \texttt{trajectory}. This records what the system produces or transforms,
  not the unit at which evidence is reported.
  \item \textbf{Automation scope}: A0--A4 as defined in
  Section~\ref{sec:framework}.
  \item \textbf{Evaluation unit}: one or more tags from
  \texttt{field\_document}, \texttt{pair}, \texttt{list}, \texttt{case},
  \texttt{trajectory}, and \texttt{outcome}. This is coded separately because,
  for example, a list-producing system can be evaluated with an outcome-level
  deployment study.
  \item \textbf{Workflow authority}: \texttt{observe}, \texttt{recommend},
  \texttt{prepare}, or \texttt{execute}, corresponding to P0--P3 in
  Figure~\ref{fig:agent-stack}. This is the highest reported workflow permission,
  not authority to make a final employment decision.
  \item \textbf{Employment-decision boundary}:\newline
  \texttt{none}, \mbox{\texttt{recommend\_only}},
  \mbox{\texttt{human\_final}},\\
  \mbox{\texttt{automated\_intermediate}},
  \mbox{\texttt{automated\_final}}, or \mbox{\texttt{not\_reported}}.
  No coded work establishes safe automated-final
  employment authority.
  \item \textbf{Pipeline stage}, \textbf{evidence source}, \textbf{learning
  signal}, and \textbf{reported evaluation design}: concise source-specific
  descriptions rather than controlled vocabularies. They preserve distinctions
  that would be lost by forcing unlike systems into one tag.
  \item \textbf{Outcome type}: \texttt{NA} or one or more controlled tags
  from\newline
  \mbox{\texttt{candidate\_response}}, \mbox{\texttt{employer\_interest}},\\
  \mbox{\texttt{marketplace\_conversion}},
  \mbox{\texttt{workflow\_uptake}},\\
  \mbox{\texttt{employment\_lifecycle}},
  \texttt{productivity}, and \mbox{\texttt{fairness\_outcome}}.
  \item \textbf{Evaluated axes}: one or more directly evaluated axes from
  \texttt{quality}, \texttt{utility}, \texttt{cost}, \texttt{fairness},
  \texttt{privacy}, and \texttt{security}. An axis discussed only as a
  limitation is not coded as evaluated.
  \item \textbf{Validity threat}, \textbf{source context}, and
  \textbf{claim boundary}: concise source-specific descriptions. The claim
  boundary states what the reported design does not license.
  \item \textbf{Publication status}: one or more tags from
  \texttt{archival}, \texttt{preprint},
  \mbox{\texttt{system\_demonstration}}, and
  \mbox{\texttt{industry\_deployment}}.
  \item \textbf{Supplementary search ID}: \texttt{NA} or an S1--S9 identifier
  linking a work to the targeted supplementary log. \texttt{NA} does not claim
  item-level provenance for the earlier primary search.
\end{enumerate}

Multiple tags within a controlled multi-value field are semicolon-separated;
\texttt{NA} denotes a documented non-applicable or unlinked value rather than an
unknown substantive result. Automation scope, evaluation unit, workflow
authority, and employment-decision boundary are interpretations of the
reported workflow, not independently dual-coded measurements. Publication
language and deployment market are interpreted separately in the synthesis,
while the companion table retains only a concise source-context field and
should not be treated as a complete regional codebook. An English-language
industry paper about a Chinese platform can still describe a Chinese deployment
context, whereas translating an English benchmark into Chinese does not
reproduce local data conventions, platform behavior, credentials, or legal
duties.

The companion manuscript bundle includes an item-level literature-map coding
file for the representative works synthesized in
Section~\ref{sec:literature-map}, together with a supplementary-search log for
the \supplementcutoff{} supplement and the \forwardcutoff{} forward update.
Section~\ref{sec:materials} gives their filenames, release identifiers, and
checksums. These files expose the coding and supplementary search used in this
synthesis, but they are not a complete historical search export and were not
independently dual-coded. Consequently, counts in them must not be interpreted
as the distribution of the full literature.

\subsection{Evidence boundaries}

Three boundaries govern our claims. First, an offline gain on private logs is
evidence about the reported task, not proof of improved hiring quality or job
performance. Second, human agreement is not automatically ground truth:
recruiters can disagree, criteria may be underspecified, and human judgments may
themselves reflect historical inequities. Third, a benchmark leaderboard is a
time-stamped observation of a particular environment. Dynamic Web content,
model updates, and tool access make agent results less stationary than
conventional test-set scores. Accordingly, we emphasize what each evaluation
can establish and what it cannot.

\section{A Unified Framework for Recruitment AI}
\label{sec:framework}

We model recruitment AI as a sequence of state transformations rather than a
single predictor. Let a request $q$ encode the hiring need, $\mathcal{C}$ the
available candidate universe, $\hat{q}$ a structured requirement profile,
$R_K=\rho(\hat{q},\mathcal{C})$ a recalled pool of at most $K$ candidates, and
$F=\phi(R_K)\subseteq R_K$ the pool that remains after explicit policy filters.
A practical system implements

\begin{equation}
q \rightarrow \hat{q} \rightarrow R_K \rightarrow F
\rightarrow L=\pi(F) \rightarrow
\{\mathcal{E}(c,q),j(c,q,\mathcal{E}),\mu(c,q,j)\}_{c\in L},
\end{equation}
where $\pi$ produces an ordered shortlist $L$, $\mathcal{E}$ is a
criterion-level evidence ledger, $j$ is a supported, contradicted, or unknown
judgment, and $\mu$ is a policy
that may shortlist, request evidence, defer, propose outreach, or execute an
authorized action. Separating $R_K$ from $F$ avoids implying that filtering
preserves pool size; separating evidence, judgment, and action prevents a
fluent rationale from silently becoming execution authority. Errors at an
early stage bound what later stages can recover. A perfect re-ranker cannot
rescue a qualified candidate absent from $R_K$, and a fluent report cannot
repair fabricated evidence.

\begin{figure}[H]
\centering
\begingroup
\renewcommand{\tiny}{\fontsize{6.9}{7.5}\selectfont}
\resizebox{0.98\textwidth}{!}{%
\begin{tikzpicture}[x=1cm,y=1cm,
  plane/.style={rounded corners=5pt, draw=black!28, fill=black!1.5,
    line width=0.8pt},
  stagecard/.style={rounded corners=5pt, draw=#1, fill=white,
    line width=1.1pt, minimum width=7.12cm, minimum height=1.34cm,
    text width=6.72cm, align=left, inner sep=5pt},
  context/.style={rounded corners=3pt, draw=stageblue!70, fill=paleblue,
    line width=0.7pt, minimum width=2.72cm, minimum height=0.52cm,
    align=center, inner sep=3pt, font=\tiny},
  authority/.style={rounded corners=3pt, draw=#1!72, fill=#1!9,
    line width=0.8pt, minimum width=3.45cm, minimum height=0.56cm,
    align=center, inner sep=2pt, font=\scriptsize\bfseries},
  controlbox/.style={rounded corners=4pt, draw=#1, fill=#1!7,
    line width=0.85pt, minimum width=3.45cm, minimum height=0.62cm,
    align=center, inner sep=3pt, font=\scriptsize\bfseries},
  outcome/.style={rounded corners=4pt, draw=#1, fill=#1!7,
    line width=0.85pt, minimum width=4.75cm, minimum height=0.86cm,
    text width=4.38cm, align=center, inner sep=3pt},
  chip/.style={rounded corners=3pt, draw=#1!70, fill=#1!9,
    line width=0.65pt, minimum width=2.94cm, minimum height=0.64cm,
    text width=2.82cm, align=center, inner sep=1pt,
    font=\fontsize{6.9}{7.5}\selectfont\bfseries, text=#1},
  directed/.style={line width=0.95pt, line cap=round, line join=round},
  flow/.style={directed,
    -{Latex[length=1.8mm,width=1.35mm]}, draw=black!62},
  controlflow/.style={directed,
    -{Latex[open,length=1.8mm,width=1.35mm]}, draw=stagepurple!82},
  feedback/.style={directed,
    -{Latex[length=1.8mm,width=1.35mm]}, draw=stagegreen!86},
  outcomeline/.style={line width=0.85pt, line cap=round,
    line join=round, draw=black!55},
  outcometap/.style={outcomeline}
]

\node[plane, draw=stagered!72, fill=palered!42,
  minimum width=15.55cm, minimum height=1.12cm] at (7.85,15.20) {};
\node[anchor=west, font=\scriptsize\bfseries, text=stagered]
  at (0.25,15.58) {GOVERNANCE ENVELOPE};
\node[chip=stagered] at (1.65,15.08) {Purpose \& consent};
\node[chip=stagered] at (4.75,15.08) {Privacy \& minimization};
\node[chip=stagered] at (7.85,15.08) {Fairness \& access};
\node[chip=stagered] at (10.95,15.08) {Security};
\node[chip=stagered] at (14.05,15.08) {Accountability};

\node[plane, fill=paleblue!24, minimum width=15.55cm,
  minimum height=1.18cm] at (7.85,13.82) {};
\node[anchor=west, font=\scriptsize\bfseries, text=stageblue]
  at (0.25,14.18) {CONTEXT INPUTS};
\node[context] (role) at (1.65,13.70)
  {\textbf{Role demand}\\criteria \textbullet\ intent};
\node[context] (candidate) at (4.75,13.70)
  {\textbf{Candidate state}\\preference \textbullet\ consent};
\node[context] (market) at (7.85,13.70)
  {\textbf{Market}\\supply \textbullet\ timing};
\node[context] (rules) at (10.95,13.70)
  {\textbf{Policy}\\rules \textbullet\ exceptions};
\node[context] (sources) at (14.05,13.70)
  {\textbf{Sources}\\ATS \textbullet\ web \textbullet\ records};

\node[plane, minimum width=15.55cm, minimum height=7.86cm]
  at (7.85,9.18) {};
\node[anchor=west, font=\scriptsize\bfseries, text=black!62]
  at (0.38,12.78) {DECISION \& EVIDENCE PLANE};

\node[stagecard=stageblue] (q) at (3.85,11.82)
  {{\scriptsize\bfseries\textcolor{stageblue}{01 \; Request model}}\\[-1pt]
   \tiny criteria \textbullet\ preferences \textbullet\ provenance\\[1pt]
   \tiny\bfseries Output: requirement schema};
\node[stagecard=stagegreen] (r) at (11.85,11.82)
  {{\scriptsize\bfseries\textcolor{stagegreen}{02 \; Recall}}\\[-1pt]
   \tiny lexical \textbullet\ dense \textbullet\ graph\\[1pt]
   \tiny\bfseries Output: candidate pool + coverage};

\node[stagecard=stagepurple] (rank) at (3.85,9.96)
  {{\scriptsize\bfseries\textcolor{stagepurple}{04 \; Re-rank}}\\[-1pt]
   \tiny fit \textbullet\ diversity \textbullet\ calibration \textbullet\ abstention\\[1pt]
   \tiny\bfseries Output: ranked list + confidence};
\node[stagecard=stageorange] (f) at (11.85,9.96)
  {{\scriptsize\bfseries\textcolor{stageorange}{03 \; Policy filter}}\\[-1pt]
   \tiny eligibility \textbullet\ consent \textbullet\ explicit policy\\[1pt]
   \tiny\bfseries Output: eligible pool + reasons};

\node[stagecard=stagecyan] (ev) at (3.85,8.10)
  {{\scriptsize\bfseries\textcolor{stagecyan}{05 \; Evidence}}\\[-1pt]
   \tiny claims \textbullet\ source spans \textbullet\ freshness \textbullet\ conflicts\\[1pt]
   \tiny\bfseries Output: criterion evidence ledger};
\node[stagecard=stagecyan] (j) at (11.85,8.10)
  {{\scriptsize\bfseries\textcolor{stagecyan}{06 \; Judgment}}\\[-1pt]
   \tiny supported \textbullet\ contradicted \textbullet\ unknown\\[1pt]
   \tiny\bfseries Output: decision record};

\node[stagecard=stagered] (a) at (3.85,6.24)
  {{\scriptsize\bfseries\textcolor{stagered}{08 \; Authorized act}}\\[-1pt]
   \tiny defer \textbullet\ ask \textbullet\ shortlist \textbullet\ contact \textbullet\ execute\\[1pt]
   \tiny\bfseries Output: external event + receipt};
\node[stagecard=stageorange] (h) at (11.85,6.24)
  {{\scriptsize\bfseries\textcolor{stageorange}{07 \; Human review}}\\[-1pt]
   \tiny inspect \textbullet\ override \textbullet\ approve \textbullet\ appeal\\[1pt]
   \tiny\bfseries Output: review / approval token};

\draw[flow] (q.east) -- (r.west);
\draw[flow] (r.south) -- (f.north);
\draw[flow] (f.west) -- (rank.east);
\draw[flow] (rank.south) -- (ev.north);
\draw[flow] (ev.east) -- (j.west);
\draw[flow] (j.south) -- (h.north);
\draw[flow] (h.west) -- (a.east);
\node[font=\tiny\bfseries, text=stagered, fill=white, inner sep=1pt] (gate)
  at (7.85,6.47) {Gate};

\node[plane, draw=stagered!45, fill=palered!18, minimum width=15.55cm,
  minimum height=0.88cm] at (7.85,4.68) {};
\node[authority=stageblue] (a0) at (1.95,4.64) {P0 \; Observe};
\node[authority=stagegreen] (a1) at (5.88,4.64) {P1 \; Recommend};
\node[authority=stageorange] (a2) at (9.82,4.64) {P2 \; Prepare};
\node[authority=stagered] (a3) at (13.75,4.64) {P3 \; Execute};
\draw[controlflow]
  (a3.north) -- ++(0,0.46) -| (gate.south);

\node[plane, fill=palepurple!18, minimum width=15.55cm,
  minimum height=0.76cm] at (7.85,3.70) {};
\node[controlbox=stageblue] (memory) at (1.95,3.68)
  {\textcolor{stageblue}{State \& memory}};
\node[controlbox=stagepurple] (controller) at (5.88,3.68)
  {\textcolor{stagepurple}{Controller}};
\node[controlbox=stagegreen] (tools) at (9.82,3.68)
  {\textcolor{stagegreen}{Tools \& perception}};
\node[controlbox=stagecyan] (trace) at (13.75,3.68)
  {\textcolor{stagecyan}{Trace \& audit}};
\draw[controlflow] (memory.north) -- (a0.south);
\draw[controlflow] (controller.north) -- (a1.south);
\draw[controlflow] (tools.north) -- (a2.south);
\draw[controlflow] (trace.north) -- (a3.south);

\node[outcome=stageblue] (employer) at (2.65,2.55)
  {{\scriptsize\bfseries\textcolor{stageblue}{EMPLOYER OUTCOMES}}\\[-1pt]
   \tiny yield \textbullet\ time \textbullet\ retention};
\node[outcome=stagegreen] (candidateout) at (7.85,2.55)
  {{\scriptsize\bfseries\textcolor{stagegreen}{CANDIDATE OUTCOMES}}\\[-1pt]
   \tiny opportunity \textbullet\ fit \textbullet\ burden};
\node[outcome=stagered] (assurance) at (13.05,2.55)
  {{\scriptsize\bfseries\textcolor{stagered}{ASSURANCE SIGNALS}}\\[-1pt]
   \tiny contest \textbullet\ harm \textbullet\ drift};
\draw[outcomeline] (2.65,3.18) -- (13.75,3.18) -- (trace.south);
\draw[outcomeline] (a.south) -- (3.85,3.18);
\draw[outcometap] (2.65,3.18) -- (employer.north);
\draw[outcometap] (7.85,3.18) -- (candidateout.north);
\draw[outcometap] (13.05,3.18) -- (assurance.north);
\draw[feedback] (assurance.east) -- ++(0.50,0) -| (15.92,14.56)
  -- (10.95,14.56) -- (rules.north);

\draw[flow] (2.30,1.48) -- (3.20,1.48);
\node[anchor=west, font=\tiny] at (3.30,1.48) {artifact / decision};
\draw[controlflow] (6.35,1.48) -- (7.25,1.48);
\node[anchor=west, font=\tiny] at (7.35,1.48) {control / permission};
\draw[feedback] (10.65,1.48) -- (11.55,1.48);
\node[anchor=west, font=\tiny] at (11.65,1.48) {outcome feedback};

\end{tikzpicture}%
}
\endgroup
\caption{Review-derived reference architecture spanning A3 adaptive observation
and optional A4 authorized execution. The complete architecture is a design
synthesis rather than a validated deployment. It separates context, decision
evidence, permission, control, and reciprocal outcomes. P0--P3 denote
permission boundaries, not the A0--A4 automation scopes in
Section~\ref{sec:framework}. Each stage exposes its output artifact, and the
authority gate prevents a fluent judgment from silently becoming an external
action. The shared gray outcome bus links authorized action and trace data to
employer, candidate, and assurance measures; green carries audited feedback.
Purple open arrows denote control or permission; red borders mark
governance or execution boundaries. Connector length reflects routing distance,
not quantitative magnitude~\citep{kenthapadi-etal-2017-linkedin,
li-etal-2025-enhancing-talent,gan-etal-2024-application,
wang-etal-2026-adaptjobrec}.}
\label{fig:agent-stack}
\end{figure}
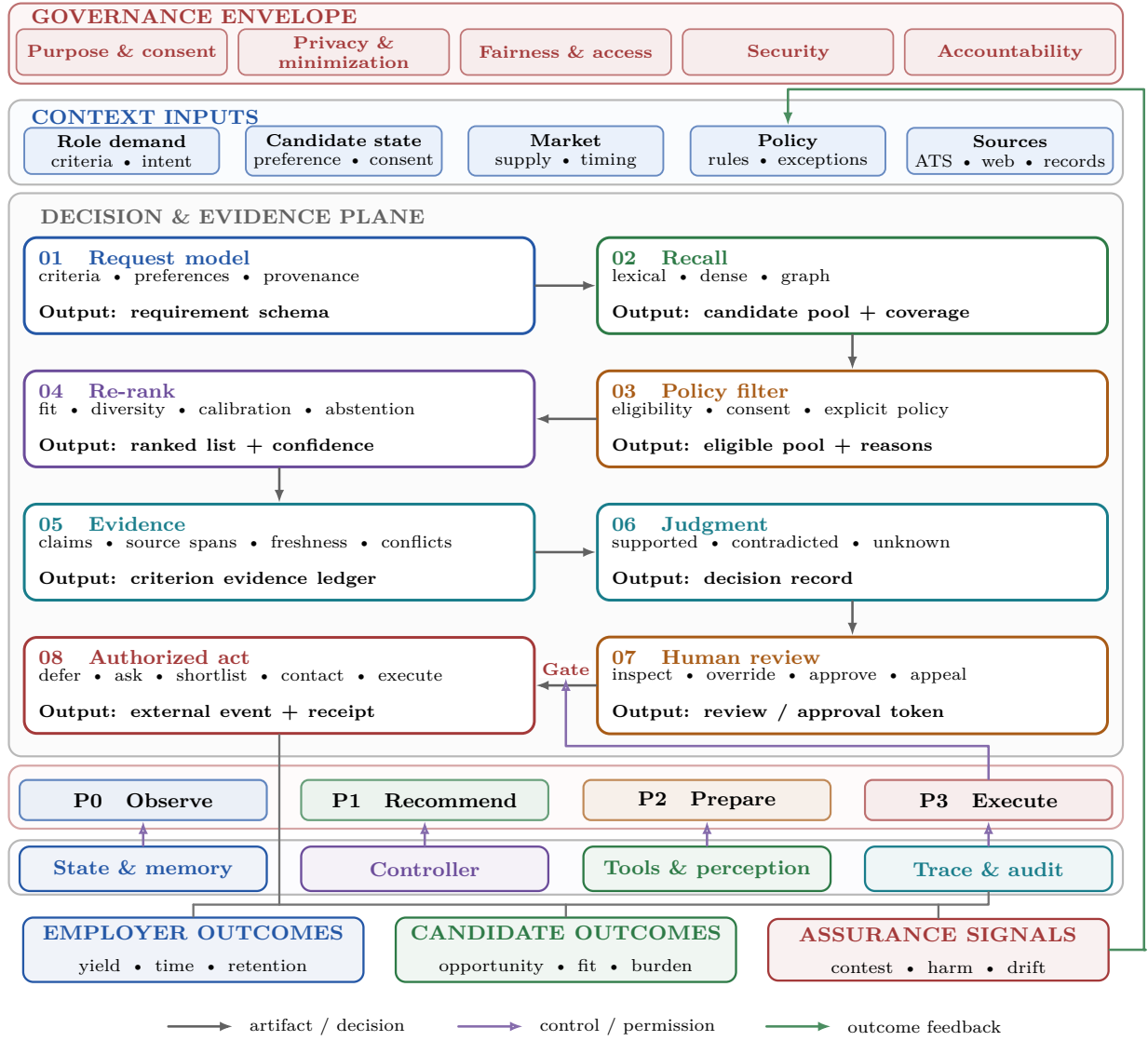

Figure~\ref{fig:agent-stack} makes the dependency structure explicit. The
execution path is surrounded by governance, observability, verification, and an
outcome loop; these controls are part of the system being evaluated rather than
documentation added after a model is chosen.

\subsection{Three axes}

\paragraph{Recruitment stage.}
The operational chain is \emph{requirement understanding $\rightarrow$
profile construction $\rightarrow$ recall $\rightarrow$ policy filtering
$\rightarrow$ re-ranking $\rightarrow$ evidence-based judgment $\rightarrow$
action $\rightarrow$ outcome}. Systems often collapse several stages, but
evaluation should preserve them.

\paragraph{Automation scope.}
We distinguish five levels: (A0) assistive transformation, such as extraction
or rewriting; (A1) predictive recommendation over a fixed corpus; (A2)
multi-stage decision support with criterion evidence and human handoff; (A3)
adaptive observation with case state, tools, and a policy that may search, ask,
or defer but cannot initiate a consequential external action; and (A4)
action-capable execution that can contact a person or mutate an external system
under explicit authority. Sourcing utility at A1--A3
does not imply selection validity. A3 requires tool-coverage, freshness, cost,
and recovery evidence; A4 additionally requires permission, confirmation,
idempotency, audit, and rollback evidence. Higher automation increases both
potential productivity and the surface for cascading errors.

Capability and granted authority are separate. We use P0--P3 for the highest
workflow permission actually reported: P0 observe, P1 recommend, P2 prepare an
external action for approval, and P3 execute it. Employment-decision authority
is coded separately because a system can conduct candidate-facing outreach or
an interview at P3 while a person retains final screening or hiring authority.
No coded work establishes safe automated-final employment authority.

\paragraph{Evaluation unit.}
The natural units are a field, pair, list, case, trajectory, and real-world
outcome. Here \emph{outcome} is the evaluation unit and \emph{lifecycle}
denotes the longitudinal observation design needed to establish it.
Pipeline dependencies link the units, but evidence at one unit does not
substitute for evidence at another.

Figure~\ref{fig:evaluation-ladder}, introduced in Section~\ref{sec:introduction},
is the paper's common claim ladder.
Section~\ref{sec:evaluation} uses L1--L6 as shorthand for the six displayed
units in the same order: field/document, pair, list, case, trajectory, and
outcome. Historical layers, mechanism labels, source types, and
validity threats annotate these units; they are not additional evaluation
hierarchies.

\subsection{Illustrative end-to-end case}

Table~\ref{tab:worked-example} applies the framework to a fictional search for
a senior retrieval engineer. It is a worked design example, not a case study,
benchmark result, or claim that the proposed controls have been validated.

\begin{table}[!t]
\centering
\footnotesize
\setlength{\tabcolsep}{4pt}
\renewcommand{\arraystretch}{1.12}
\caption{Illustrative end-to-end evaluation trace; no empirical results are reported.}
\label{tab:worked-example}
\begin{tabularx}{\textwidth}{@{}P{2.15cm}P{6.25cm}Y@{}}
\toprule
\textbf{Checkpoint} & \textbf{Illustrative system state and exposed failure} &
\textbf{Evidence gate} \\
\midrule
Request / profile & ``Senior retrieval engineer'' is decomposed into
production retrieval experience, preferred multilingual work, and an unresolved
location constraint. The failure is silently converting a preference or
ambiguity into a hard exclusion. &
Reviewer-approved hard/preferred criteria, unresolved questions, and a versioned
request. \\
Document / recall & Resume claims retain source spans; lexical, dense, and graph
channels construct a pool. OCR errors, rare job titles, or one missing channel
can make a qualified person unreachable. &
Critical-field accuracy and provenance, full-corpus Recall@$K$, channel
ablation, and coverage by language and seniority. \\
Filter / rank & Work authorization remains \emph{unknown} until supported;
feasible candidates are ranked under a fixed rubric. The failure is treating
missing information as negative or hiding a policy filter inside one score. &
Filter trace and override policy, nDCG@$K$, qualified yield, pool-composition
stability, and slice results. \\
Case judgment & Each criterion is marked supported, contradicted, or unknown
with cited evidence. The failure is a persuasive recommendation whose decisive
claim has no source. &
Evidence precision and recall, expert adjudication, calibration, and
risk--coverage under selective deferral. \\
Handoff / action & The system proposes an outreach message but cannot send it
until an authorized reviewer sees the pool, filters, evidence, and uncertainty.
The failure is decorative approval or an irreversible action without authority. &
Correction and override quality, action log, permission tests, latency, cost,
and recovery from rejected actions. \\
Lifecycle & Contact, interview, offer, acceptance, start, performance, and
retention are followed from information available at each checkpoint. The
failure is scoring with future information or coding unobserved outcomes as
negative. &
Eligible risk set, time-stamped features, follow-up window, delayed labels,
censoring, adverse impact, and a stated causal estimand. \\
\bottomrule
\end{tabularx}
\end{table}

\subsection{Reciprocity and partial observability}

Recruitment is reciprocal: a candidate may prefer a job for which they are not
qualified, while an employer may prefer a candidate who will not accept. This
distinguishes job markets from conventional user--item recommendation
\citep{malinowski-etal-2006-bilateral,palomares-etal-2020-reciprocal}. Moreover,
resumes and job descriptions are incomplete views of latent states. Missing
evidence should remain \emph{unknown}, not be silently converted into a negative
fact. Let

\begin{equation}
\begin{aligned}
p_P(c,q)
  &=P(\text{candidate is willing to pursue }q\mid q,c,I_C),\\
p_Q(c,q)
  &=P(\text{candidate satisfies job-related criteria}
      \mid q,c,\mathcal{E}(c,q),I_E),
\end{aligned}
\end{equation}
where $I_C$ and $I_E$ are the information available to the candidate and
employer at the decision time, $p_P$ denotes candidate-side preference, and
$p_Q$ denotes qualification against job-related criteria. A reciprocal policy
uses both probabilities, their uncertainty, evidence-acquisition cost, contact
burden, and applicable constraints to choose whether to search, ask, defer,
recommend, or propose contact. It should report candidate-side exposure,
preference satisfaction, unwanted-contact burden, correction or contest
outcomes, and eventual acceptance alongside employer-side qualified yield.
Offer acceptance is a downstream outcome observed only for the eligible risk
set; it is not interchangeable with the preference probability $p_P$. This
separation becomes essential when an agent can acquire evidence or ask a
clarifying question instead of forcing an immediate match.

\section{Technical Evolution: From Search and Ranking to Agents}
\label{sec:evolution}

The five historical layers below overlap; ``era'' names a readable lineage,
not a mutually exclusive code or a claim to cover every earlier hiring
technology. Production systems usually retain elements from several layers
because high-throughput retrieval, hard policy constraints, and human approval
remain useful even when an LLM is added.

\subsection{Era I: rules, ontologies, and bilateral recommendation}

Within the search-and-recommender lineage reviewed here, early electronic
recruitment systems emphasized Boolean retrieval, manually curated fields,
taxonomies, and content-based similarity
\citep{de-ruijt-bhulai-2021-review,mashayekhi-etal-2024-challenge}. These methods
made requirements inspectable but were brittle to synonyms, implicit skills,
cross-lingual variation, and inconsistent resume structure. The bilateral
recommendation formulation of \citet{malinowski-etal-2006-bilateral} was an
important conceptual advance: matching people to jobs requires modeling two
selection processes, not merely recommending an inanimate item to a user.

\citet{paparrizos-etal-2011-machine} formulated next-job prediction as
supervised learning over public career transitions. This reframed job search as
a behavioral prediction problem but also foreshadowed a persistent limitation:
observed transitions reveal what happened under a particular labor market and
exposure process, not all mutually beneficial matches that could have happened.
Skill normalization systems later combined extraction, entity normalization,
and occupational taxonomies to support search and labor-market analytics at
scale~\citep{javed-etal-2017-skill}. Such symbolic resources remain valuable for
hard constraints, cross-document normalization, and auditability.

\subsection{Era II: industrial retrieval, behavioral learning, and marketplaces}

As recruitment platforms accumulated clicks, applications, messages, and
hires, the architecture converged with search and advertising systems. A
request is served by multiple retrieval channels, candidates are filtered by
availability or policy, and increasingly expensive rankers operate on a
shrinking pool. LinkedIn's production paper describes an online multi-tier
serving architecture backed by offline model updates
\citep{kenthapadi-etal-2017-linkedin}. Similar systems combine profile content,
context, activity, and marketplace constraints. Talent-search work further
documents query understanding, retrieval, personalization, and entity-aware
ranking as distinct serving components~\citep{geyik-etal-2018-talent}.

This era established three lessons that remain central. First, recall and
ranking are different problems. Second, objectives are multi-sided: sending
too many applicants to one job can burden employers and reduce each applicant's
chance, even when click-through rises. Third, platform feedback is
selection-biased. A missing application can mean lack of exposure, lack of
interest, perceived low probability of success, or true incompatibility.

Learning-to-rank and multi-task models can optimize several behaviors, but each
label sits at a different point in the funnel. Clicks are frequent and weak;
offers or successful hires are sparse and delayed. Industrial papers reporting
online conversion gains therefore provide strong evidence for marketplace
utility under their deployment conditions, while not by themselves validating
the substantive quality of selection decisions.

The 2017 ACM RecSys Challenge on XING made cold-start deployment unusually
visible: teams first passed a fixed historical evaluation and then submitted
recommendations that were served in a live push-recommendation setting
\citep{xing-2017-recsys-challenge}. The target combined candidate interest with
employer-side appropriateness. This is stronger operational evidence than a
single offline split, but the competition outcome still does not establish
long-term employee performance or transportability beyond that marketplace.

\subsection{Era III: neural semantic matching and structured representations}

Neural models reduced dependence on exact term overlap. PJFNN learned joint
hierarchical representations of job requirements and resume experiences from
historical applications~\citep{zhu-etal-2018-pjfnn}. APJFNN added ability-aware
attention to expose which experiences contributed to individual requirements
\citep{qin-etal-2018-apjfnn}. These models made person--job fit an end-to-end
text learning problem and supplied limited interpretability through alignment
weights.

Transferable global matching addressed cross-domain reuse
~\citep{bian-etal-2019-domain}, while context-aware transformers represented
resume sections and competence levels under an expert annotation protocol
~\citep{li-etal-2020-competence}. These studies broadened the supervision
design beyond raw application labels, although both remained bounded by their
occupational and platform contexts.

Subsequent work targeted the weaknesses of behavioral supervision. A
multi-view co-teaching network combined text and relation views to reduce the
effect of sparse, noisy interactions~\citep{bian-etal-2020-coteaching}.
DPGNN then modeled the two directions explicitly by representing candidate and
employer selection preferences in a dual-perspective interaction graph,
including successful and failed matches~\citep{yang-etal-2022-dpgnn}. Its three
real-world recruitment datasets show the value of reciprocal structure, while
remaining an offline evaluation of platform-specific behavioral labels.
Knowledge-graph and career-path models represented higher-order relationships
and temporal development rather than treating a resume as a bag of sentences
\citep{gong-etal-2024-career}. DuerQuiz connected skill extraction, a skill
graph, and personalized question recommendation in a deployed interview
assessment workflow~\citep{qin-etal-2019-duerquiz}.

JobHop v2 subsequently released 355,315 quarter-indexed career trajectories
extracted from roughly 440,000 pseudonymized multilingual resumes, and STEP
used temporal and educational signals for next-job prediction
\citep{johary-etal-2026-jobhop,johary-etal-2026-step}. These resources improve
trajectory modeling and reproducibility, but an observed or predicted next job
is not the same as evaluating a recruitment policy across offer, start,
performance, and retention.

Transformer encoders and contrastive learning strengthened semantic retrieval.
ConFit augments selected resume and job sections and exploits in-batch negatives
\citep{yu-etal-2024-confit}; ConFit v2 generates a hypothetical reference resume
for each job and mines runner-up hard negatives from unlabeled data
\citep{yu-etal-2025-confit}. JobFormer combines skill-aware structure with
semantic-enhanced transformer representations~\citep{guan-etal-2025-jobformer}.
The methodological shift is important: performance increasingly depends on
which positives, negatives, and intermediate structures are constructed, not
only on encoder capacity.

PJB adds a fixed full-corpus retrieval protocol with complete job descriptions,
nearly 200,000 de-identified resumes across six domain families, and diagnostic
labels for job-competency reasoning~\citep{wang-etal-2026-pjb}. Its domain and
reasoning slices expose failure modes hidden by aggregate scores, while its
positive-only judgments and internally sourced corpus limit claims about
complete recall and external transportability.

The era did not solve validity. Attention is not proof, historical applications
remain confounded, and many datasets are private. Pairwise AUC or classification
accuracy can improve while the full-corpus shortlist remains unchanged.

\subsection{Retrieval systems require stage-level disclosure}

The label ``matching model'' hides choices that determine whether a result is
reachable in practice. Table~\ref{tab:retrieval-disclosure} compares the public
disclosures of representative systems. NR means not reported in the cited
public source; it does not mean the capability is absent from production.

\begingroup
\scriptsize
\setlength{\tabcolsep}{2.4pt}
\renewcommand{\arraystretch}{1.12}
\begin{longtable}{@{}P{1.85cm}P{2.8cm}P{2.7cm}P{2.35cm}P{3.1cm}P{2.75cm}@{}}
\caption{Stage-level retrieval and ranking disclosure in representative work.}
\label{tab:retrieval-disclosure}\\
\toprule
\textbf{Work} & \textbf{Candidate generation / pool} &
\textbf{Negatives or supervision} & \textbf{Re-ranking} &
\textbf{Offline and online evidence} & \textbf{Latency and index update} \\
\midrule
\endfirsthead
\multicolumn{6}{l}{\textit{Table~\ref{tab:retrieval-disclosure} continued}}\\
\toprule
\textbf{Work} & \textbf{Candidate generation / pool} &
\textbf{Negatives or supervision} & \textbf{Re-ranking} &
\textbf{Offline and online evidence} & \textbf{Latency and index update} \\
\midrule
\endhead
\bottomrule
\endlastfoot

LinkedIn ranking &
Multi-tier retrieval over a production corpus
~\citep{kenthapadi-etal-2017-linkedin}; stage pool sizes NR &
Behavioral and profile signals; negative construction NR &
Progressively more expensive ranking stages &
Production architecture and operational lessons; complete-pool Recall@$K$ NR &
Offline model updates and serving constraints described; exact latency and
freshness NR \\

XING RecSys 2017 &
Team-defined generation over XING users for new-job targets
~\citep{xing-2017-recsys-challenge}; per-target pool NR &
Candidate response plus employer-side appropriateness under organizer protocol &
Team-specific ranking before submission &
Fixed offline gate followed by live online response; Recall@$K$ was not the
primary live criterion &
Targets changed daily; serving latency and index-refresh mechanics NR \\

DPGNN &
Candidate--job interaction graph bounded by each dataset
~\citep{yang-etal-2022-dpgnn} &
Successful and failed matches; quadruple ranking and dual-perspective
contrastive losses &
Reciprocal graph score; no separate production re-ranker reported &
Offline ranking on three real recruitment datasets; no online test &
Serving latency, incremental graph update, and index freshness NR \\

ConFit family~\citep{yu-etal-2024-confit,yu-etal-2025-confit,
yu-etal-2026-confit-v3} &
Dense dual-encoder retrieval over dataset corpora; v3 re-ranks a retrieved
top-20 pool &
In-batch and rejected pairs; v2 mines runner-up negatives around the top
3--4\% rather than treating the top results as negatives &
Cosine retrieval followed by multi-pass listwise LLM re-ranking in v3 &
nDCG@10 and recall-based offline tests; no live deployment metric &
Distillation addresses inference cost; serving latency and index-refresh
policy NR \\

PJB~\citep{wang-etal-2026-pjb} &
Fixed full-corpus evaluation with nearly 300 job descriptions and nearly
200,000 de-identified resumes across six domain families &
More than 2,000 positive job-competency judgments; missing pairs may be
unjudged rather than negative &
Separate query-rewriting and depth-20 re-ranking tracks &
nDCG@10, Recall@20, MRR@10, domain-family, and reasoning-type diagnostics; no
online experiment &
Offline corpus; serving latency and index-update policy NR \\

Role-aware talent search~\citep{li-etal-2025-enhancing-talent} &
Search-engine recall and pre-ranking over a private candidate index; pool size NR &
Click-through rate (CTR), conversion rate (CVR), and resume relevance &
Role-aware mixture-of-experts final ranker with LLM-derived job features &
Offline AUC plus online A/B test; 17.29\% lift in click-through conversion rate &
Production serving is described; exact latency and index-update interval NR \\

Jobindex &
Private CV database and recommender; retrieval-pool mechanics NR
~\citep{kaya-bogers-2026-jobindex} &
Model-development signals are reported, but negative construction is not the
focus of the deployment analysis &
Suggested slate reviewed by recruiters &
Adoption, engagement, and task success over 17 months and 41,390 jobs &
Serving latency and candidate-index refresh NR \\

Open-web agents~\citep{chen-etal-2025-xbench,
shi-etal-2026-peoplesearchbench} &
Search tools over a non-enumerable live Web; no fixed complete candidate pool &
Criterion verification and task rubrics rather than training negatives &
Agent filtering and shortlist synthesis after search &
Task completion and qualified-result yield; Recall@$K$ against a complete
universe is not identifiable &
Steps and elapsed work can be logged; source and index freshness are externally
controlled \\

\end{longtable}
\endgroup

The missing cells are consequential. Without the candidate universe and stage
pool sizes, pair accuracy cannot be translated into reachable recall. Without
negative semantics, offline ranking gains are hard to interpret. Without
latency, refresh cadence, and online metrics, a model comparison does not yet
specify a deployable retrieval system.

\subsection{Era IV: LLM-augmented recruiting pipelines}

Beginning in 2023, LLMs appeared in four recurring roles.

\paragraph{Data generator and editor.}
GIRL generates an ideal job description using supervised fine-tuning, a reward
model, and reinforcement learning before grounding the representation to
available jobs~\citep{zheng-etal-2026-girl}. Other work uses LLMs to complete
sparse resumes and adversarially align low- and high-quality representations
\citep{du-etal-2024-job-gan}. These approaches can expose latent preference or
improve weak text, but they also risk converting plausible model knowledge into
false biographical evidence.

\paragraph{Feature engineer and teacher.}
Industrial systems use LLMs to extract fine-grained requirements or rewrite
low-quality job descriptions, while smaller mixture-of-experts or ranking
models serve high-volume traffic~\citep{li-etal-2025-enhancing-talent,
chen-etal-2026-category}. This division
of labor often offers a better latency--quality trade-off than invoking a large
generative model for every pair.

\paragraph{Re-ranker and comparator.}
LLMs can compare multiple candidates, generate criterion-specific rationales,
and create listwise supervision. ConFit v3 explores multi-pass listwise
re-ranking, teacher distillation, data cleaning, and reinforcement learning
\citep{yu-etal-2026-confit-v3}. The relevant baseline is therefore no longer a
single zero-shot prompt: it is a cascade with strong retrieval, calibrated
listwise comparison, and cost controls.

\paragraph{Document interpreter.}
ResumeBench demonstrates that multilingual, multi-layout resume parsing remains
difficult even when models produce schema-valid JSON
\citep{ling-etal-2025-beyond}. \mbox{JobResQA} moves from fields to evidence-grounded
questions over resumes and job descriptions~\citep{carrino-etal-2026-jobresqa}.
These tasks show why upstream document errors must remain visible in downstream
evaluation.

LLM augmentation improves linguistic coverage and supervision flexibility, but
introduces evaluator dependence, model-version drift, and circularity. If the
same model family generates, filters, and judges examples, apparent gains may
reflect stylistic affinity rather than better recruitment decisions.

\subsection{Era V: tool-using recruiting agents}

Agentic systems change the execution boundary. Instead of scoring a supplied
pair, an agent may clarify a role, search databases and the public web, merge
entities, reconstruct a candidate's career, compare evidence against a rubric,
draft outreach, and log a handoff. A modular screening framework separates
classification, summarization, scoring, and decision stages
\citep{gan-etal-2024-application}. A newer HR agent integrates job descriptions,
CVs, interview transcripts, and recruiter feedback, and uses an active listwise
tournament for candidate comparison~\citep{yuksel-etal-2026-agentic}.
Requirement-driven sourcing makes an earlier dependency explicit: TalentScout
elicits, validates, retrieves, and verifies an under-specified people request,
while TalentTrace evaluates 21 systems over 691 requirements
\citep{he-etal-2026-requirement-sourcing}. Its union-pooled, multi-model oracle
improves open-world comparability but does not make complete recall identifiable
or independent of the evaluated system family.

Work described as agentic spans several interaction patterns. MockLLM simulates
interviewer and candidate roles to generate additional bilateral evidence
\citep{sun-etal-2024-mockllm}; AdaptJobRec selects recommendation tools and
varies planning depth with query complexity
\citep{wang-etal-2026-adaptjobrec}; and SimRPD uses a user simulator to select
training conversations for proactive recruitment dialogue
\citep{cao-etal-2026-simrpd}. These systems should not be collapsed into one
capability claim merely because they contain multiple modules or LLM roles.

The distinction between an LLM and an agent is operational rather than
rhetorical. An agent has state, tools, a policy over actions, and an environment
that can change during execution. This creates new failure modes: search can
miss a source, entity resolution can merge two people, a page can become stale,
an outreach action can occur without approval, or repeated searches can consume
unbounded cost. The appropriate evaluation object is the trajectory, including
intermediate evidence and actions, not only the final report.

Table~\ref{tab:agent-types} separates four patterns often described as
agentic. Static modular workflows and role simulation are adjacent control
patterns rather than agents under our operational definition. Adaptive
tool-use systems meet the definition at scope A3; only action-capable systems
at A4 can initiate consequential external actions.

\begin{table}[H]
\centering
\footnotesize
\setlength{\tabcolsep}{3.2pt}
\renewcommand{\arraystretch}{1.12}
\caption{Agent classes by state, tool use, workflow authority, and
employment-decision boundary.}
\label{tab:agent-types}
\begin{tabularx}{\textwidth}{@{}P{2.5cm}P{3.1cm}P{3.2cm}P{3.0cm}Y@{}}
\toprule
\textbf{Class / scope} & \textbf{State and control flow} & \textbf{Tool or environment} &
\textbf{Representative work} & \textbf{Required evidence boundary} \\
\midrule
Static modular workflow / A2 & Fixed modules or rubric stages; case state passes
between components & Supplied documents; no necessary environmental
observation or mutation & Modular screening and HR
assessment~\citep{gan-etal-2024-application,yuksel-etal-2026-agentic} &
Module errors, calibration, and human review \\
Role simulation / adjacent to A2 & Dialogue state among simulated roles &
Generated interview or recruiting conversations &
MockLLM~\citep{sun-etal-2024-mockllm} & Simulator validity and transfer to real
users \\
Adaptive tool use / A3 & Planner chooses searches, recommendation tools, or
reasoning depth & Read-only or bounded APIs; Web state may change &
AdaptJobRec, TalentScout, xbench, and PeopleSearchBench
~\citep{wang-etal-2026-adaptjobrec,he-etal-2026-requirement-sourcing,
chen-etal-2025-xbench,shi-etal-2026-peoplesearchbench} & Tool coverage,
evidence freshness, recovery,
cost, and judge replacement \\
Action-capable agent / A4 & Policy can initiate consequential external actions &
Candidate-facing dialogue or interview administration; external systems may
change & SimRPD and AI voice interviews
~\citep{cao-etal-2026-simrpd,jabarian-henkel-2026-voice-ai} & Candidate-facing
execution does not imply automated employment decisions; require explicit
authorization, audit, recovery, and human control of final decisions \\
\bottomrule
\end{tabularx}
\end{table}

Static modular workflows and role simulation can provide useful evidence
without satisfying the operational definition of an agent. A3 adds adaptive,
stateful observation of an environment but no authority to initiate a
consequential external action. A4 adds candidate-facing or system-mutating
execution and therefore requires separate evidence about permissions,
irreversible effects, human approval, and recovery. SimRPD executes proactive
candidate dialogue, and the voice-AI field experiment executes interview
administration; neither result establishes automated-final employment authority.

Figure~\ref{fig:evolution} summarizes the five historical layers. A production
system may combine lexical retrieval, behavioral supervision, a neural ranker,
an LLM evidence module, and an agentic handoff. Comparative analysis therefore
uses the three primary axes in Section~\ref{sec:framework}; model
representation, evidence source, supervision signal, market sidedness,
decision authority, context, and validity threat are secondary annotations
described in Section~\ref{sec:protocol}. The historical labels remain narrative
waypoints rather than a second taxonomy.

\section{Capabilities Across the Recruitment Workflow}
\label{sec:workflows}

The same model can appear in several workflow stages, but the evidence required
to justify its use changes with the decision being supported.

\subsection{Requirement understanding and document intelligence}

A recruitment request need not arrive as a clean query. Job descriptions mix mandatory
requirements, preferences, boilerplate, and organizational context. Resumes mix
explicit facts, omissions, layout cues, and self-presentation. Systems therefore
need to extract entities, normalize skills, distinguish hard from soft
constraints, and preserve provenance.

Large-scale skill normalization established a hybrid pattern in which named
entity recognition feeds a canonical occupational inventory
\citep{javed-etal-2017-skill}. Recent LLM systems make schema creation and
cross-lingual extraction easier, but ResumeBench shows that format compliance is
not semantic correctness~\citep{ling-etal-2025-beyond}. The safest intermediate
representation is not a flat candidate profile. It is a set of typed claims
linked to document spans and carrying one of at least three states:
\emph{supported}, \emph{contradicted}, or \emph{unknown}. This prevents missing
evidence from being silently interpreted as failure.

Interactive sourcing research makes requirement elicitation itself measurable:
an agent can ask bounded clarifying questions, validate acceptance criteria,
and preserve a two-stage commitment before retrieving people
\citep{he-etal-2026-requirement-sourcing}. This is useful evidence about request
formation and sourcing breadth, not proof that the resulting criteria are
job-related or lawful.

For research, upstream evaluation should include field-level correctness,
hierarchy, document localization, and critical-error severity. For deployment,
it should also measure how extraction errors alter recall and decisions. A
parser with slightly lower average F1 may be preferable if its residual errors
rarely delete mandatory qualifications.

\subsection{Candidate and job retrieval}

Retrieval answers a coverage question: did the system place potentially relevant
items inside a budgeted pool? Content, collaborative, graph, and dense channels
often have complementary failure modes. Lexical retrieval is strong for exact
credentials and rare technologies; dense retrieval captures paraphrases;
behavioral channels capture revealed interest; graph channels expose related
organizations or career transitions. Hybrid fusion is therefore common.

Knowledge graphs can also make skill and occupation relationships explicit.
LLM-enhanced graph representations have been studied for online job
recommendation~\citep{wu-etal-2024-graph-job}, and JobMatchAI combines a graph,
semantic retrieval, and explanations in a system demonstration
~\citep{vyas-etal-2026-jobmatchai}. Such combinations improve inspectability,
but a graph edge still needs a defined source and update policy.

The main methodological risk is evaluating only on observed positive pairs.
People who never saw a job cannot express interest, and candidates outside a
platform database cannot be retrieved at all. Recall should be reported against
an independently judged or otherwise defensible target set, with separate
coverage by occupation, seniority, language, and data-source availability.

PeopleSearchBench illustrates an emerging evaluation pattern for open talent
search: queries are decomposed into explicit criteria, returned people are
checked against live external evidence, and systems are scored on relevance,
qualified-result yield, and information utility
\citep{shi-etal-2026-peoplesearchbench}. This is closer to a recruiter's
information task than fixed pair classification, although live-web verification
introduces temporal and source-availability dependencies.

\subsection{Filtering, re-ranking, and candidate assessment}

Filtering enforces policy or feasibility constraints before ranking. Examples
include work authorization, location, salary bands, candidate consent, or
minimum credentials. These constraints should not be buried in an opaque score:
their state, source, and override policy need explicit representation.

Re-ranking orders the feasible pool under a particular objective. Pointwise
scoring is cheap but poorly calibrated across candidates; pairwise comparison
captures relative preference; listwise methods better represent the delivered
shortlist but face context length and order effects. LLM listwise re-ranking can
use a role-specific rubric and surface reasons, but explanations must be
evaluated against source evidence rather than plausibility.

Assessment is a stronger claim than ranking. The observational comparison of
GPT-4 and recruiter ratings on real resumes finds limited agreement and
dimension-specific differences; prompting improves some correlations but does
not make the judgments interchangeable~\citep{vaishampayan-etal-2025-human}.
Controlled validity tests similarly distinguish ranking candidates with known
qualification differences from abstaining when candidates are substantively
equivalent~\citep{castleman-etal-2026-validity}. These results support a
selective-decision interface: recommend when evidence and separation are
sufficient; otherwise request a fact, show a tie, or defer.

\subsection{Interviewing, communication, and candidate support}

AI can generate interview questions, conduct practice interviews, summarize
transcripts, and draft outreach. DuerQuiz is an early deployed example that
connects job-skill knowledge to personalized written assessment
\citep{qin-etal-2019-duerquiz}. LLMs make adaptive dialogue easier, yet dialogue
quality and selection validity remain different constructs. A conversation can
be fluent while asking irrelevant questions, overstepping consent, or rewarding
communication style unrelated to job performance.

Candidate experience is not merely a usability outcome. Organizational-justice
theory treats job relatedness, opportunity to perform, consistency, explanation,
and respectful treatment as part of how applicants evaluate a selection
procedure~\citep{gilliland-1993-fairness}. Across 86 independent samples,
applicant perceptions were associated with organizational attractiveness,
offer-acceptance intentions, and recommendation intentions
~\citep{hausknecht-etal-2004-applicant}. Studies of highly automated interviews
found differences in perceived control, social presence, fairness, and
acceptance in their tested settings, and examined two-way communication as one
possible explanatory pathway
~\citep{langer-etal-2019-automated,acikgoz-etal-2020-justice}. These reactions
are consequential, but neither favorable perceptions nor acceptance prove
predictive validity, legal compliance, or equal outcomes.

A large natural field experiment adds unusually direct outcome evidence. The
firm received 70,884 applications, of which 67,056 were eligible and randomized
among human-interviewer, AI-interviewer, and interviewer-choice conditions. The
headline estimates compare the 53,660 applications assigned directly to the AI
and human conditions: the AI condition increased offers, starts, and
short-horizon retention, while human recruiters retained evaluation and hiring
authority~\citep{jabarian-henkel-2026-voice-ai}. The productivity result is
narrower. Only a subset of hires had performance records, and the three measures
used different worker-month panels; within those observed subsets, the authors
reported no statistically significant or economically meaningful performance
differences. The evidence is further bounded to one recruitment provider,
predominantly entry-level service roles, applicant refusal, and technical
failures; it does not validate autonomous selection across occupations.

Candidate response also depends on who visibly owns a consequential decision.
Two mixed-method studies found that adding human-manager review after an
AI-based rejection improved applicants' reactions to the organization
\citep{zhou-etal-2026-rejected}. This supports substantive review rather than a
decorative approval button, but it does not establish that the reviewed decision
was more accurate or fair.

Evaluation should therefore separate: coverage of job-relevant competencies;
factual grounding in the candidate's materials; consistency across equivalent
responses; accommodation for disability or language variation; candidate
experience; and whether any generated score predicts a defensible criterion.
The objective for outreach should not be reply rate alone. Misleading
personalization, sensitive inference, excessive contact, and unapproved external
actions create harms even if conversion improves.

Recent dialogue systems widen this stage from question generation to adaptive
multi-turn execution. MockLLM uses simulated roles to augment matching evidence
~\citep{sun-etal-2024-mockllm}, whereas SimRPD uses simulator-based data
evaluation and selection to train candidate-facing dialogue toward contact
acquisition in a real-world recruitment scenario
~\citep{cao-etal-2026-simrpd}. The latter is bounded P3 workflow execution, not
authority to make an employment decision. Both evaluations should be read
against the dialogue goal they optimize, not as interchangeable evidence of
hiring validity.

Candidate-facing systems deserve equal attention. They can explain requirements,
identify evidence gaps, compare jobs, or help draft materials. However, the
same generative assistance can homogenize applications and create a feedback
loop in which evaluators prefer model-written text. Decision systems should not
confuse writing polish with qualification, and evaluation should include
semantic-equivalence tests across writing styles.
Controlled hiring-pipeline experiments also show that prompt guardrails and
human checkpoints reduce different forms of unsupported resume content but
neither intervention removes every subtle error~\citep{takano-2026-fabrication}.

\subsection{End-to-end agents and human handoff}

An end-to-end recruiting agent is best understood as a policy over tools and
handoffs. A minimal safe trajectory contains:

\begin{enumerate}
  \item a structured request with unresolved ambiguities;
  \item a source-bounded search plan;
  \item entity-resolved candidates with timestamps;
  \item criterion-level evidence and uncertainty;
  \item a shortlist or comparison under an explicit objective;
  \item proposed actions, separated from executed actions; and
  \item a human-readable audit record.
\end{enumerate}

AdaptJobRec provides a concrete example of conditional orchestration: simple
queries use direct tool selection, while complex queries invoke memory
filtering and task decomposition~\citep{wang-etal-2026-adaptjobrec}. This design
turns latency into a routing problem rather than assuming that every request
needs the deepest reasoning path.

xbench's recruitment tasks make company mapping, information retrieval, and
talent sourcing observable as professional work rather than isolated QA
\citep{chen-etal-2025-xbench}. This is an important step toward measuring
utility. Still, a professional benchmark and an employment-selection validation
study answer different questions. Completing a sourcing task does not establish
that the resulting screening or hiring decision is job-related, fair, lawful,
or causally beneficial.

\section{Representative Literature and Evidence Map}
\label{sec:literature-map}

Table~\ref{tab:literature-map} is the main evidence map for all 40 coded works.
It retains each source-specific evaluated object and claim boundary rather than
placing industrial online effects, public benchmark scores, and controlled
validity audits on one undifferentiated leaderboard. A compact visual topology
of the same landscape is provided in
Appendix~\ref{app:evidence-landscape}.

The table is deliberately selective rather than a paper-count table. Its status
markers distinguish archival peer-reviewed work (\textsuperscript{A}),
preprints (\textsuperscript{P}), system-demonstration papers
(\textsuperscript{D}), and studies with reported live industry or deployment
evidence (\textsuperscript{I}); markers may be combined.

Evidence maturity is kept separate from topical novelty. Archival studies and
production evaluations can support results within their reported settings;
preprints, benchmarks, and system demonstrations are retained chiefly when they
expose a new artifact or evaluation problem. In particular, the 2025--2026
agent rows establish an emerging research surface, not a mature evidence base
for autonomous employment decisions. Status markers in the Work column make
that boundary visible without converting source type into a numerical quality
score.

For compactness, the table has 37 display rows: PJFNN and APJFNN share one row,
and the three ConFit versions share another. The companion CSV preserves all 40
atomic records. Within this selected set, the coded automation scopes are
A0=2, A1=24, A2=8, A3=4, and A4=2; workflow authority is P0=3, P1=27, P2=8,
and P3=2. These are descriptions of the purposive map, not field-prevalence
estimates. The employment-decision boundary remains a separate per-item field.

\begingroup
\footnotesize
\setlength{\tabcolsep}{3.5pt}
\renewcommand{\arraystretch}{1.13}
\begin{longtable}{@{}P{3.05cm}P{3.15cm}P{4.15cm}P{4.75cm}@{}}
\caption{Representative recruitment-AI literature in 37 display rows
representing 40 atomic coded works. Status: \textsuperscript{A} archival
peer-reviewed; \textsuperscript{P} preprint; \textsuperscript{D} system
demonstration; \textsuperscript{I} live industry or deployment evidence.
``Private'' means the underlying data are not a generally reusable public
benchmark; it does not imply weak execution.}
\label{tab:literature-map}\\
\toprule
\textbf{Work} & \textbf{Stage / object} & \textbf{Reported evidence} &
\textbf{Claim boundary} \\
\midrule
\endfirsthead
\multicolumn{4}{l}{\textit{Table~\ref{tab:literature-map} continued}}\\
\toprule
\textbf{Work} & \textbf{Stage / object} & \textbf{Reported evidence} &
\textbf{Claim boundary} \\
\midrule
\endhead
\midrule
\multicolumn{4}{r}{\textit{Continued on next page}}\\
\endfoot
\bottomrule
\endlastfoot

Bilateral recommendation\textsuperscript{A}\newline
\citep{malinowski-etal-2006-bilateral} &
Person--job recommendation &
Two-sided preference formulation and prototype comparison &
Conceptual and offline evidence; no modern full-corpus or deployment audit \\

Machine-learned job recommendation\textsuperscript{A}\newline
\citep{paparrizos-etal-2011-machine} &
Next-job prediction &
Supervised prediction over observed career transitions &
Observed transitions conflate exposure, opportunity, and preference \\

LinkedIn job recommendation\textsuperscript{A,I}\newline
\citep{kenthapadi-etal-2017-linkedin} &
Industrial retrieval and ranking &
Production architecture and operational lessons &
Platform utility is deployment-specific; selection validity is outside scope \\

XING RecSys Challenge\textsuperscript{I}~\citep{xing-2017-recsys-challenge} &
Cold-start push recommendation &
Historical gate followed by live online evaluation &
Marketplace response is stronger than offline-only evidence but not long-term
employment validity \\

PJFNN / APJFNN\textsuperscript{A}~\citep{zhu-etal-2018-pjfnn,qin-etal-2018-apjfnn} &
Pair matching and alignment &
Offline classification / ranking on application data &
Historical applications are behavioral labels; attention is not proof \\

Fair talent ranking\textsuperscript{A,I}~\citep{geyik-etal-2019-fairness} &
Top-$K$ re-ranking &
Utility under target representation constraints; production application &
Target distribution is a policy choice, not a complete fairness definition \\

Domain adaptation\textsuperscript{A}~\citep{bian-etal-2019-domain} &
Cross-domain person--job fit &
Transferable representations on private recruitment datasets &
Transfer gains do not resolve label ambiguity or external validity \\

Competence-level matching\textsuperscript{A}~\citep{li-etal-2020-competence} &
Resume level and pair classification &
Expert-annotated domain data, inter-rater agreement, classification accuracy &
Single occupational context; pair accuracy does not measure search recall \\

Multi-view co-teaching\textsuperscript{A}~\citep{bian-etal-2020-coteaching} &
Sparse interaction matching &
Offline gains from text and relation views &
Noise robustness is evaluated within historical platform data \\

DPGNN\textsuperscript{A}~\citep{yang-etal-2022-dpgnn} &
Reciprocal graph matching &
Directed candidate and employer preferences on three real-world datasets &
Two-sided modeling remains offline and label validity is platform-specific \\

Embedding rank attacks\textsuperscript{A}~\citep{samadi-etal-2021-attacks} &
Adversarial resume ranking &
White- and black-box attacks against TF--IDF and sentence embeddings &
Attack study uses simplified rankers; production defenses remain open \\

ConFit family\textsuperscript{A,P}~\citep{yu-etal-2024-confit,
yu-etal-2025-confit,yu-etal-2026-confit-v3} &
Dense recall and LLM re-ranking &
Contrastive retrieval, hard negatives, listwise teacher / re-ranker tests &
Benchmark gains depend on constructed positives, negatives, and judges \\

PJB\textsuperscript{P}~\citep{wang-etal-2026-pjb} &
Full-corpus person--job retrieval &
Nearly 200,000 resumes across six domains with competency and reasoning
diagnostics &
Positive-only judgments and an internally sourced corpus limit complete-recall
and external-validity claims \\

Graph understanding with LLMs\textsuperscript{A}~\citep{wu-etal-2024-graph-job} &
Job-graph recommendation &
LLM-enhanced graph representations and offline recommendation metrics &
Graph data and offline labels do not establish end-to-end workflow value \\

Job-GAN\textsuperscript{A}~\citep{du-etal-2024-job-gan} &
Resume enrichment and recommendation &
Generative completion plus adversarial representation learning &
Plausible completion can introduce unsupported biographical content \\

MockLLM\textsuperscript{P}~\citep{sun-etal-2024-mockllm} &
Simulated interview and bilateral assessment &
Role-play dialogue quality and person--job matching performance &
Synthetic interviews are added evidence, not validated selection instruments \\

Modular LLM screening workflow\textsuperscript{A}~\citep{gan-etal-2024-application} &
Case-level screening workflow &
Classification, summarization, scoring, and decision modules &
Framework evaluation is not a labor-market outcome or legal-validity study \\

Attribute or Abstain\textsuperscript{A}~\citep{buchmann-etal-2024-attribute} &
Long-document grounded assistance &
Attribution correctness and abstention behavior &
General document benchmark; recruitment transfer must be tested \\

ResumeBench\textsuperscript{A}~\citep{ling-etal-2025-beyond} &
Multilingual resume parsing &
Multi-layout schema extraction and critical field errors &
Document correctness alone does not establish shortlist quality \\

Role-aware talent search\textsuperscript{A,I}~\citep{li-etal-2025-enhancing-talent} &
Industrial candidate ranking &
Private offline metrics and online A/B tests for click-through rate (CTR) / conversion &
Strong marketplace evidence under one workflow; causal hiring quality remains open \\

Jobindex human-in-the-loop recommender\textsuperscript{A,I}\newline
\citep{kaya-bogers-2026-jobindex} &
Recruiter-facing candidate recommendation &
Seventeen-month deployment spanning 41,390 jobs, adoption, and task success &
Workflow uptake does not establish long-term job performance or general fairness \\

xbench\textsuperscript{P}~\citep{chen-etal-2025-xbench} &
Professional recruiting trajectory &
Dynamic company mapping, research, and talent-sourcing tasks &
Task completion does not validate employment-selection decisions \\

JobMatchAI\textsuperscript{A,D}\newline
\citep{vyas-etal-2026-jobmatchai} &
Knowledge-graph matching platform &
System demonstration combining graph, semantic search, and explanation &
Demonstration evidence is not a controlled comparative outcome study \\

Multi-document HR assessment\textsuperscript{A,D}\newline
\citep{yuksel-etal-2026-agentic} &
Multi-document candidate assessment &
Rubric modules and active listwise tournament in a system demo &
Case utility and predictive validity require independent expert evaluation \\

AdaptJobRec\textsuperscript{A}\newline
\citep{wang-etal-2026-adaptjobrec} &
Conversational job-recommendation agent &
Real recommendation scenarios and accuracy/latency ablations &
Conversation utility does not establish employer-side qualification validity \\

SimRPD\textsuperscript{A,I}~\citep{cao-etal-2026-simrpd} &
Candidate-facing proactive dialogue &
Simulator-based data selection and real-scenario candidate communication &
P3 dialogue execution does not confer employment-decision authority; goal
completion and conversion require consent and experience audits \\

PeopleSearch\allowbreak Bench\textsuperscript{P}\newline
\citep{shi-etal-2026-peoplesearchbench} &
Open-web people search &
Criterion verification, qualified-result yield, and information utility &
Live-web scores depend on time, source access, and incomplete ground truth \\

Resume-screening validity\textsuperscript{P}~\citep{castleman-etal-2026-validity} &
Ranking and abstention &
Controlled qualification differences and equivalence cases &
Construct-focused audit; not a full deployed hiring-system study \\

Prompt injection in resumes\textsuperscript{A}~\citep{baxi-etal-2026-prompt} &
Document-to-ranking attack &
Controlled influence of embedded instructions on LLM screening &
Attack success does not by itself identify an effective production defense \\

External personality stability audit\textsuperscript{A}
~\citep{rhea-etal-2022-stability} &
Case and repeated-measure assessment audit &
Ninety-four participants and two commercial systems tested across source,
file-format, embedded-URL, and temporal perturbations &
Instability is necessary-condition evidence, not criterion validity or a
deployed selection effect \\

LLM trait inference\textsuperscript{A}~\citep{hartel-2026-tasks-traits} &
Case-level assessment &
Assessment-center transcripts, repeated LLM scores, and a human-rater benchmark &
Reliability and above-chance convergence remain construct- and input-dependent \\

Conditional resume-bias audit\textsuperscript{A}~\citep{pavlopoulos-2026-minimal} &
Pair and case audit &
30,384 controlled evaluations with positive controls, equivalence tests, and
prompt stress tests &
Small effects in synthetic resumes do not certify deployed thresholds or models \\

Hiring-pipeline fabrication\textsuperscript{P}~\citep{takano-2026-fabrication} &
Document and case workflow &
180 controlled runs comparing automation, prompt guardrails, and a human checkpoint &
Small synthetic setting; neither mitigation removed every subtle unsupported claim \\

Requirement-driven sourcing\textsuperscript{P}
~\citep{he-etal-2026-requirement-sourcing} &
Request and sourcing trajectory &
691 requirements evaluated across 21 systems with elicitation and verification &
Union-pooled candidates and model judges approximate rather than enumerate open-world recall \\

AI voice interviews\textsuperscript{P,I}\newline
\citep{jabarian-henkel-2026-voice-ai} &
P3 interview administration and outcome &
70,884 applications received; 67,056 eligible applications randomized, with
offer, start, and retention outcomes &
Humans retained final decisions; productivity was observed only in
measure-specific worker subsets, and one service setting bounds transportability \\

End-to-end public-employment audit\textsuperscript{P,I}
~\citep{galdon-clavell-2026-applied-filtered} &
Pipeline and outcome audit &
Approximately 497,000 records across seven human and automated stages &
Missing qualification and later employer outcomes prevent causal attribution \\

LongPIBench\textsuperscript{P}~\citep{liu-etal-2026-longpibench} &
Long-document security &
Synthetic and real-world long-context prompt-injection tests, including resume screening &
Multi-domain benchmark results do not measure production attack prevalence \\

\end{longtable}
\endgroup

Within this purposively selected map, pair- and list-level technical evidence is
more developed than evidence about trajectory safety, human correction, or
long-term outcomes. This is a comparison inside the selected set, not an
estimate of field-wide prevalence. Recent agent benchmarks broaden task scope
faster than they establish employment-validity claims; useful future
comparisons should therefore connect several evaluation units rather than
replace pair metrics with one end-to-end score.

Table~\ref{tab:quantitative-snapshot} preserves several reported numbers without
treating unlike tasks, denominators, or outcomes as a common effect size. A
performance delta, deployment sample, and benchmark scale answer different
questions; each row therefore retains its comparator, uncertainty, and claim
boundary.

\begin{table}[H]
\centering
\scriptsize
\setlength{\tabcolsep}{3.2pt}
\renewcommand{\arraystretch}{1.12}
\caption{Representative quantitative evidence snapshot. Values use each
source's reported metric and denominator and are not pooled or directly
comparable.}
\label{tab:quantitative-snapshot}
\begin{tabularx}{\textwidth}{@{}P{2.6cm}P{3.5cm}P{4.75cm}Y@{}}
\toprule
\textbf{Work} & \textbf{Setting} & \textbf{Reported quantitative evidence} &
\textbf{Comparator, uncertainty, and claim boundary} \\
\midrule
Role-aware talent search~\citep{li-etal-2025-enhancing-talent} &
Private production ranking logs and a nine-working-day online test &
Relative AUC gains of 1.70\% for CTR and 5.97\% for CVR; online click-through
conversion rate increased 17.29\%. &
GDCN online baseline with a 1:1 traffic split; reported $p=0.01598$.
Supports marketplace behavior under that workflow, not hiring quality. \\
ConFit v2~\citep{yu-etal-2025-confit} &
Resume and job ranking on two real-world datasets &
Compared with ConFit, average absolute improvements reported as 13.8\% in
recall and 17.5\% in nDCG across ranking directions and datasets. &
Non-deterministic methods were averaged over three runs; no live deployment
metric or employment-validity outcome. \\
ResumeBench~\citep{ling-etal-2025-beyond} &
Multilingual, multi-layout resume parsing benchmark &
2,500 synthetic resumes, 50 templates, 30 career fields, five languages, and
24 evaluated LLMs. &
Quantifies benchmark breadth rather than a hiring effect; synthetic generation
and layout/language coverage bound transfer. \\
Jobindex recommender~\citep{kaya-bogers-2026-jobindex} &
Recruiter-facing human-in-the-loop deployment &
Seventeen months of use covering 41,390 jobs, with adoption, engagement, and
task-success analyses. &
Longitudinal workflow evidence, but the study does not yield one transferable
causal lift for job performance or fairness. \\
xbench recruitment~\citep{chen-etal-2025-xbench} &
Dynamic profession-aligned agent benchmark &
50 real-world headhunting tasks spanning company mapping, information
retrieval, and talent sourcing. &
Establishes task coverage and initial agent baselines; scores remain dependent
on environment, judge, and benchmark version. \\
PeopleSearch\allowbreak Bench~\citep{shi-etal-2026-peoplesearchbench} &
Four platforms on 119 multilingual queries across recruiting, sales, expert
search, and influencer discovery &
The leading system scored 65.2 overall, 18.5\% above the runner-up, and
completed all 119 queries; verification agreement reported $\kappa=0.84$. &
Confidence intervals and human validation are reported, but live-web access,
time, and incomplete ground truth limit transportability. \\
Requirement-driven sourcing~\citep{he-etal-2026-requirement-sourcing} &
Interactive candidate sourcing over live-web evidence &
691 requirements and 21 systems; the reported system covered all requirements
and returned 2.5 times the yield of the comparison set. &
Relevance and recall depend on a union-pooled candidate set and multi-model
oracle; the system authors also participate in the comparison. \\
AI voice interviews~\citep{jabarian-henkel-2026-voice-ai} &
70,884 applications received; 67,056 eligible applications randomized; the
headline AI-versus-human comparison covers 53,660 direct assignments &
The AI-interview condition increased the offer rate by 12\% relative to the
human-interviewer condition; starts and
at-least-one-month retention increased 18\%. In the observed worker-performance
subsets, no statistically significant or economically meaningful productivity
difference was reported. &
Humans retained final evaluation and hiring authority; productivity records do
not cover all hires or randomized applications. One provider, labor market, and
job family limit transportability, and some applicants refused or experienced
technical failure. \\
Public-employment fairness audit~\citep{galdon-clavell-2026-applied-filtered} &
Seven-stage semi-automated pipeline, 2017--2022 &
Approximately 497,000 candidate--vacancy records; aggregate parity masked
salary-, age-, and intersection-specific disparities. &
Observational audit with missing qualification and later employer outcomes;
vendor, human, data, and pipeline effects cannot be separated causally. \\
\bottomrule
\end{tabularx}
\end{table}

\section{Industrial Evidence and Regional Deployment Context}
\label{sec:industrial-regional}

Private training data and confidential experimentation prevent any public
survey from reconstructing a production recruiting system completely. Public
evidence can nevertheless be made more useful by triangulating sources that
answer different questions. Table~\ref{tab:industrial-evidence} separates five
such layers. The key discipline is to retain the claim boundary: a production
paper can support an architecture or online metric, an annual report can
support strategic and geographic scope, and a regulator audit can support an
observed compliance failure. None alone establishes causal hiring quality.

\begingroup
\small
\setlength{\tabcolsep}{4pt}
\renewcommand{\arraystretch}{1.13}
\begin{longtable}{@{}P{2.75cm}P{4.2cm}P{4.55cm}P{4.15cm}@{}}
\caption{What public industrial evidence can and cannot establish.}
\label{tab:industrial-evidence}\\
\toprule
\textbf{Evidence layer} & \textbf{Illustrative primary evidence} &
\textbf{What it adds} & \textbf{Persistent blind spot} \\
\midrule
\endfirsthead
\multicolumn{4}{l}{\textit{Table~\ref{tab:industrial-evidence} continued}}\\
\toprule
\textbf{Evidence layer} & \textbf{Illustrative primary evidence} &
\textbf{What it adds} & \textbf{Persistent blind spot} \\
\midrule
\endhead
\bottomrule
\endlastfoot

Industry paper & LinkedIn talent search describes multi-pass retrieval,
real-time updates, recruiter feedback, reciprocal interest, and production
constraints~\citep{geyik-etal-2018-talent}; role-aware talent ranking reports
private offline and online tests~\citep{li-etal-2025-enhancing-talent}. &
System architecture, objective choice, evaluation protocol, and sometimes
online effect under a real workflow. &
Private data, experiment allocation, costs, negative results, and long-term
employment outcomes are usually unavailable. \\

Engineering or product disclosure & Indeed describes AI-supported candidate
summaries, outreach, and sourcing, with self-reported usage and conversion
statistics~\citep{indeed-2024-smart-sourcing}; an official algorithm-filing
disclosure describes 51job's use of resume, job, search, exposure, click, and
application signals for deep-learning-based prediction, ranking, diversification,
and real-time feedback~\citep{qianjin-nd-51job-recommendation}. &
Product surface, operational sequence, user scale, and the business metric the
company chooses to disclose. &
Marketing selection, non-public denominators, and lack of independent causal
or fairness evaluation limit inference. \\

Annual report & SEEK reports more than 750 million daily employment-activity
data points, over 60 million candidate profiles, over 16 million verified
credentials, and certification of its AI-governance management system
~\citep{seek-2026-annual-report}. KANZHUN describes two-sided recommendation,
its recruitment-focused Nanbeige LLM, and enterprise AI functions at different
stages of application, gray-scale testing, or exploration
~\citep{kanzhun-2026-annual-report}. Tongdao Liepin reports candidate-search,
job-seeker, and headhunter agents spanning requirement understanding, sourcing,
communication, and interview support~\citep{tongdao-liepin-2026-interim}. &
Board-level strategic commitment, operating geography, governance statements,
and organizational investment. &
Model design, slice performance, candidate-level error, and causal contribution
to reported financial outcomes remain opaque. \\

Regulator audit or engagement & The UK Information Commissioner's Office audited recruitment
AI providers, issued almost 300 recommendations, and reported issues including
protected-attribute filters, inferred ethnicity or gender, over-collection,
and indefinite retention~\citep{ico-2024-recruitment-audit}. Its later
Recruitment Rewired report synthesizes voluntary engagement with more than 30
employers~\citep{ico-2026-recruitment-rewired}. &
Observed deployment practices, concrete compliance failures, remediation, and
procurement questions external to vendor self-report. &
Provider identities can be anonymized, and voluntary employer engagement is
not an audit or investigation; neither source compares predictive quality or
generalizes automatically to other jurisdictions. \\

Public algorithmic-transparency record & The UK Department for Education
describes a pre-deployment apprenticeship-vacancy QA API with 11 checks,
risk-tiered human sampling, per-check error metrics, and fail-to-human behavior
~\citep{uk-dfe-2026-vacancy-qa}. &
Named owner, workflow phase, models, inputs, outputs, routing thresholds,
historical evaluation, risks, mitigations, and operational fallback. &
The record is unusually detailed but remains a pre-deployment self-disclosure;
rare positive cases, human-derived labels, and undisclosed denominators limit
inference from reported perfect sensitivity on one check. \\

\end{longtable}
\endgroup

Product changes can also reveal an authority boundary without establishing why
it changed. Indeed paused the automatic-application mode of Apply For Me and
continued the mode in which job seekers review drafts before submission
\citep{indeed-2026-apply-for-me}. This is a documented move from execution to
preparation plus confirmation; the announcement attributes it to feedback and
does not establish a safety incident, regulatory intervention, or causal quality
failure.

\subsection{Evidence from non-English and locally regulated markets}

Geographic coverage is not publication language. An English-language paper
about a Chinese platform can describe a Chinese deployment, while a translated
benchmark does not reproduce local credentials, labor-market practice,
population, product role, or law. Table~\ref{tab:regional-rules} therefore maps
selected primary rules and regulator evidence to concrete system controls. It
is a time-stamped technical comparison, not legal advice, a product-specific
applicability decision, or a global census. The selected jurisdictions expose
different control boundaries; omitted markets remain open transportability
questions rather than implied absences.

The targeted Chinese-language supplement also found locally situated evidence
that is easy to miss in an English-only search. A qualitative study combined
social-media posts from people reporting AI-interview experience with
interviews of 22 applicants, and developed a grounded account linking system
and content design to perceived fairness and acceptance
~\citep{liu-etal-2023-ai-interview}.
Chinese HR-decision experiments associated algorithmic decision-making with
lower perceived transparency and procedural fairness in the studied scenarios
~\citep{pei-etal-2021-procedural-fairness}, while a recent Chinese-language
review organizes applicant, organizational, and technical threats to perceived
fairness in AI recruitment~\citep{han-chen-2025-ai-recruitment}. These sources
do not estimate prevalence across China, but they supply local constructs and,
in the applicant studies, candidate perspectives that platform metrics and
translated benchmarks omit.

A 2026 China--Pakistan study extends this candidate perspective with 225 and
213 respondents who reported recent AI-video-interview experience. Perceived AI
interview experience was negatively associated with transparency, interactive
communication, and job-pursuit intention, with trust moderating the associations
\citep{zhao-etal-2026-applicant-perceptions}. The convenience samples and
retrospective measures included no randomized human-interview condition, so the
results are not causal country comparisons.

Local evidence also supplies two more concrete evaluation paths. A Chinese job-
advertisement gender lexicon, developed from 53,786 advertisements, can support
pre-publication wording detection and auditing
~\citep{jiang-etal-2023-chinese-gender-lexicon}; it does not show that a detected
or rewritten advertisement changes hiring outcomes. Internal data from a
Chinese job board were used to estimate that explicit gender requests shifted
the composition of applicant pools and were associated with callback penalties for
gender-mismatched applicants~\citep{kuhn-etal-2020-gender-targeted}. That study
is not an algorithm audit, but it demonstrates why exposure, self-selection,
application, callback, and hiring should be evaluated as distinct lifecycle
stages.

\begingroup
\footnotesize
\setlength{\tabcolsep}{3.5pt}
\renewcommand{\arraystretch}{1.14}
\begin{longtable}{@{}P{2.15cm}P{4.35cm}P{5.15cm}P{3.95cm}@{}}
\caption{Selected regional rules and recruitment-system implications.
Legal status checked \legalcutoff.}
\label{tab:regional-rules}\\
\toprule
\textbf{Region} & \textbf{Primary source, actor, and scope} &
\textbf{Concrete design implication} & \textbf{Current status and boundary} \\
\midrule
\endfirsthead
\multicolumn{4}{l}{\textit{Table~\ref{tab:regional-rules} continued}}\\
\toprule
\textbf{Region} & \textbf{Primary source, actor, and scope} &
\textbf{Concrete design implication} & \textbf{Current status and boundary} \\
\midrule
\endhead
\bottomrule
\endlastfoot

China & PIPL Article~19 generally requires the shortest retention period
necessary for the processing purpose unless law or administrative regulation
provides otherwise. Article~24 requires automated decision-making to be
transparent, fair, and impartial; when a decision has a significant impact on
personal rights or interests, the individual may request an explanation and
refuse a decision made solely through automated processing
~\citep{china-2021-pipl}. Article 26 of the Network Recruitment
Service Management Provisions requires platform-based human-resource service
providers to keep recruitment and service information for at least three years
after service completion~\citep{mohrss-2020-network-recruitment}. &
Classify records before assigning retention: candidate personal information,
recruitment/service transaction records, model logs, and audit evidence may
have different purposes and legal bases. Preserve explicit filter and ranking
logic and correction channels, and support explanation and refusal rights where
an automated decision significantly affects personal rights or interests. &
Both rules are in force, but the three-year duty should not be read as a blanket
retention period for every item of candidate personal data. Applicability also
depends on product role; public-facing algorithm and generative-AI services
have additional scope rules~\citep{china-2022-algorithm-rules,
china-2023-genai-measures}. \\

China: platform recruitment information & A five-department notice issued on
25 December 2025 requires online platforms to verify recruitment-service
accounts and recruitment information, prohibit algorithmic amplification of
false recruitment information, establish risk-identification models, preserve
evidence, and provide rapid complaint handling and human review for concentrated
complaints~\citep{china-2025-platform-recruitment-notice}. &
Implement account and job provenance checks, content-integrity gates,
amplification blocks, auditable incident records, and complaint-triggered human
review before treating a listing as a valid retrieval target. &
The notice is in force and was published online in January 2026. Its central
scope is the authenticity of recruitment information and platform-account
governance; it is not a complete rule for candidate-ranking fairness or
selection validity. \\

European Union & Specified recruitment and candidate-selection uses appear in
Annex~III and are classified as high-risk through Article~6(2). Article~6(3)
can exclude an Annex~III system that poses no significant risk and satisfies a
listed condition, but the derogation never applies when the system performs
profiling of natural persons~\citep{eu-2024-ai-act}; GDPR Article~22 separately
covers specified solely automated decisions with legal or similarly significant
effects~\citep{eu-2016-gdpr}. Article~50 covers specified directly interactive
AI systems~\citep{ec-2026-article50-guidelines}, while the new Article~4a
creates a narrowly conditioned route for bias detection or correction using
special-category data~\citep{eu-2026-ai-omnibus}. &
Allocate actor-specific controls: provider duties include system compliance,
documentation, conformity, registration, corrective action, and post-market
monitoring; deployer duties include instructed use, competent human oversight,
operational monitoring, controlled logs, and applicable notices. Isolate
protected-attribute audit data behind necessity, pseudonymization, access,
deletion, and logging controls. &
Regulation (EU) 2026/1744 was published in the Official Journal on 24 July
2026 and entered into force on 27 July 2026. It amended Article~113 so that
Chapter~III, Sections~1--3, except Article~6(5), apply from 2 December 2027 to
systems classified as high-risk under Article~6(2) and Annex~III, including
covered recruitment systems. The 2 August 2028 date applies instead to the
Article~6(1) and Annex~I product branch~\citep{eu-2026-ai-omnibus}. GDPR duties
already apply, and Article~50 has applied since 2 August 2026. Article~4a does
not itself require bias detection or authorize protected attributes as ordinary
ranking features. \\

United Kingdom & The ICO recruitment-AI audit applies UK data-protection duties
to sourcing, screening, and selection tools~\citep{ico-2024-recruitment-audit}.
Section 80 of the Data (Use and Access) Act 2025 (DUAA) replaced UK GDPR Article
22 with Articles 22A--22D~\citep{uk-2025-duaa}. &
Ask vendors for a defined purpose, data-protection roles, retention periods,
candidate notices, data-minimization evidence, discrimination monitoring, and
proof that inferred attributes are necessary and accurate. For solely automated
significant decisions, provide information, a route to representations and
challenge, and human intervention. &
Section~80 and the related Schedule~6 provisions commenced on 5 February 2026.
Under the saving provision in SI 2026/82, the amendments apply to decisions
taken on or after that date; earlier decisions remain subject to the saved prior
rules~\citep{uk-2026-duaa-commencement}. The DUAA broadens use for
non-special-category data while retaining stricter conditions for
special-category data. The UK and EU regimes should not be treated as identical. \\

United States / New York City & Under the Uniform Guidelines Q\&A, validation is
desirable generally, but the Guidelines require validity evidence when a
selection procedure has adverse impact~\citep{eeoc-1979-uniform-guidelines-qa}.
NYC Local Law 144 requires a covered automated
employment decision tool to receive a recent bias audit, publish a summary, and
provide notices~\citep{nyc-2021-local-law-144,nyc-dcwp-2023-aedt}. &
Map each score or filter to a job-related construct, preserve validation and
adverse-impact evidence, support reasonable accommodation, and implement a
jurisdiction-aware audit and notice release gate before use. &
The Q\&A is interpretive agency guidance, not itself binding law
\citep{eeoc-2020-guidance-nonbinding}; NYC's rule is in force. Underlying
statutes, regulations, and case-specific facts determine legal obligations,
which can arise without a Uniform Guidelines validation trigger. NYC's
definition and audit metrics do not cover every AI-assisted workflow. \\

\end{longtable}
\endgroup

This comparison changes the external-validity question. The relevant unit is
not ``English versus non-English papers,'' but a tuple of language, labor-market
practice, platform population, product role, and applicable law. A credible
multi-region study should report all five and should avoid extrapolating from one
platform's behavior labels or one jurisdiction's audit definition to another.

\section{Evaluation: From Offline Accuracy to Professional Utility}
\label{sec:evaluation}

Evaluation is the central methodological bottleneck because recruitment systems
optimize multiple outcomes under incomplete counterfactuals.
Figure~\ref{fig:evaluation-ladder} supplies the common unit-to-claim ladder:
evidence at a lower unit does not answer a higher-level workflow, validity, or
business question. This section operationalizes the ladder for consequential
selection, labels, retrieval, grounding, human collaboration, agent tasks, and
outcome evidence.

\subsection{Personnel-selection validity is a use argument}

When a system moves from sourcing assistance to consequential selection, model
accuracy is only one part of validity. Job analysis must first define the work,
the relevant knowledge, skills, abilities, and other characteristics, and the
criterion the procedure is intended to predict. Reliability or repeatability
is necessary for a score to be interpretable, but a consistently reproduced
score can still measure the wrong construct.

Traditional selection practice distinguishes content evidence, criterion-related
evidence, and construct evidence; the appropriate mix depends on the intended
use~\citep{eeoc-1979-uniform-guidelines-qa}. Validation is not a fixed property
of a score in isolation, but an argument connecting job constructs,
operational measures, criteria, and the personnel decision in which the score
is used~\citep{binning-barrett-1989-validity}. A defensible study should therefore
connect score content to job requirements, test prediction against future
job-relevant criteria rather than recruiter agreement alone, and examine
differential prediction and subgroup error. It should also report incremental
validity and operational utility beyond a strong existing procedure---for
example, a structured interview, work sample, or production ranker---rather
than only a weak prompt baseline. Structured interviews illustrate how
standardized questions, scoring, and job relatedness strengthen an assessment
process~\citep{campion-etal-1997-structured-interview}; recent corrections to
selection-validity meta-analysis also caution against treating inherited
validity coefficients as fixed constants~\citep{sackett-etal-2022-validity}.

Recent evidence separates three quantities that are often collapsed. In an
assessment-center study, repeated LLM trait scores were reasonably consistent,
but convergence with established trait measures was modest and varied with the
behavioral information available~\citep{hartel-2026-tasks-traits}. More
generally, a correlation between an AI score and subject-matter-expert ratings
is not automatically inter-rater reliability; factor-analytic estimation is
needed when the measures are not parallel~\citep{speer-2026-ai-irr}. Finally,
near-zero aggregate score changes after ChatGPT's release provide evidence
against large population-wide inflation in the studied operational tests, but
cannot identify individual AI use or rule out construct and criterion drift
\citep{dabdoub-pool-2026-testing}.

SIOP's recommendations make the boundary explicit: AI-based selection
assessments should face the same scrutiny for validity, consistency, job
relevance, fairness, and documentation as traditional instruments
\citep{siop-2023-ai-assessment}. These requirements do not make every sourcing
recommendation a psychological test. They become increasingly relevant as a
tool filters, scores, assesses, or materially determines access to employment.

A black-box stability audit exposes an even more basic prerequisite. Across 94
participants, two commercial personality-prediction systems changed materially
across input source, file format, embedded-URL, or time perturbations
~\citep{rhea-etal-2022-stability}. Stability is necessary but not sufficient
for validity: the study used a narrow technical-graduate-student sample and did
not test downstream hiring outcomes.

\subsection{Labels are constructs, not facts}

Clicks, applications, recruiter advances, interviews, offers, and accepted jobs
represent different constructs. For an outcome $Y_s$ observable only after
exposure $E_s$ and a preceding action $A_s$, the observed funnel probability
factorizes as

\begin{equation}
P(Y_s=1\mid c,q) =
P(E_s=1\mid c,q)\,
P(A_s=1\mid E_s=1,c,q)\,
P(Y_s=1\mid A_s=1,E_s=1,c,q).
\end{equation}

This identity uses a sequential funnel-gating condition
$Y_s=1 \Rightarrow A_s=1 \Rightarrow E_s=1$. It describes the probability of
an observed positive event; it does not license coding every downstream
non-observation as a substantive negative. When $E_s=0$ or $A_s=0$, the later
potential outcome is unobserved and should be marked structurally missing or
censored rather than treated as $Y_s=0$.

Each factor is affected by rank position, platform design, brand, labor-market
conditions, and prior decisions. This observed funnel probability is not the
latent suitability $Z$ of a pair. Training directly on $Y_s$ can be useful for
predicting the same stage, but it does not identify $Z$ without additional
assumptions. Negative sampling makes the problem sharper: an unobserved pair is
not necessarily an unsuitable pair. Hard-negative mining
\citep{yu-etal-2025-confit} improves learning efficiency, yet the semantics of
the negative pool still require explanation.

Assessment integrity is also part of label validity. In a study combining 219
professionals with generated response sets from three frontier models,
forced-choice personality formats did not uniformly prevent human or
AI-assisted faking and did not clearly improve convergent or criterion-related
validity~\citep{robie-etal-2026-faking}. Format choice is therefore one control,
not proof that the observed score retains its intended meaning.

Expert labels have their own uncertainty. A strong protocol defines the rubric,
collects multiple ratings, records disagreement, and uses adjudication without
erasing the original distribution. Agreement with a single recruiter should not
be called human-level performance unless the variability among recruiters is
also reported.

\subsection{Retrieval and ranking}

For nonnegative graded relevance labels $rel_i$, nDCG@$K$ evaluates ordering:

\begin{equation}
\mathrm{nDCG}@K = \frac{\sum_{i=1}^{K}
(2^{rel_i}-1)/\log_2(i+1)}{\mathrm{IDCG}@K},
\end{equation}
with nDCG defined as zero when the ideal discounted cumulative gain is zero.
Its interpretation depends on the completeness and independence of the
judgments. In recruitment, there may be many acceptable candidates, only a few
of whom appear in historical data. Pooling, expert review of a fixed top-$K$,
or criterion-grounded verification can supply more defensible judgments. A
complete report should include recall, precision or nDCG, qualified-result
yield, coverage across slices, and the size and provenance of the candidate
universe.

Ranking should also be stress-tested under semantically equivalent document
changes, candidate order permutations, and different pool compositions. If a
candidate's rank changes materially when unrelated candidates are added, the
system may be unsuitable for high-stakes comparison even when average nDCG is
strong.

\subsection{Grounding, uncertainty, and selective prediction}

LLM outputs require claim-level evidence. Citation correctness asks whether a
source supports a claim; citation completeness asks whether material claims are
covered. General grounded-generation benchmarks show why fluent answers and
well-formed citations are insufficient~\citep{gao-etal-2023-enabling,
niu-etal-2024-ragtruth}. Fine-grained diagnostics separate retrieval from
generation errors~\citep{ru-etal-2024-ragchecker}, while evaluator frameworks
make judge calibration an explicit part of measurement
~\citep{saad-falcon-etal-2024-ares}. Recruiting adds typed uncertainty: a
missing degree, date, or work authorization is not equivalent to contradictory
evidence.

Selective prediction evaluates a system as the acceptance threshold varies. If
$r_i$ is a severity-weighted loss and $g_i(\tau)$ indicates acceptance at
threshold $\tau$, then, for $\sum_i g_i(\tau)>0$,

\begin{equation}
R_{\mathrm{sel}}(\tau) =
\frac{\sum_i g_i(\tau)r_i}{\sum_i g_i(\tau)}, \qquad
C(\tau) = \frac{1}{n}\sum_i g_i(\tau).
\end{equation}

The zero-coverage endpoint is reported separately rather than assigned an
arbitrary risk. Risk--coverage curves reveal whether a model can route
uncertain cases instead of forcing a score for every candidate; review time and
evidence-acquisition cost should be reported alongside the curve rather than
hidden inside average loss. A recruitment-specific policy should distinguish
global refusal from a targeted action such as ``request proof of
certification'' or ``confirm current location.'' Long-document attribution and
selective-refusal benchmarks provide reusable methods for this interface
\citep{buchmann-etal-2024-attribute,muhamed-etal-2026-refusalbench}. Their
transfer to recruitment must retain criterion-level consequences: abstaining
on a missing phone number is not equivalent to abstaining on a mandatory
license.

\subsection{Human--AI evaluation}

Human review is not a cost-free oracle. The interface can anchor reviewers on a
model score, make unsupported reasons appear authoritative, or shift effort
toward confirming rather than independently assessing evidence. Observed
human--LLM disagreement~\citep{vaishampayan-etal-2025-human} should therefore be
studied with randomized interfaces, blind baselines, decision time, correction
rate, and error severity. A useful experiment compares at least:

\begin{enumerate}
  \item human alone;
  \item AI prediction without rationale;
  \item AI prediction with an evidence-linked rationale; and
  \item a selective system that defers or requests information.
\end{enumerate}

Outcomes should include both average quality and differential reliance: do
reviewers defer to the system more for some groups, job families, or confidence
levels? A nominal human-in-the-loop design is weak evidence unless humans have
time, authority, and information to correct the system.

Deployment evidence now goes beyond conceptual interface proposals. The
Jobindex candidate recommender was studied over 17 months and 41,390 jobs,
including recruiter adoption, trust, engagement with suggested candidates, and
task success~\citep{kaya-bogers-2026-jobindex}. This is direct evidence about a
human--algorithm workflow. It does not, by itself, establish long-term job
performance, transportability, or the absence of differential harm.

The randomized AI-voice interview study goes farther along the funnel by
reporting offers, starts, retention, and productivity after automating interview
administration while leaving evaluation and hiring with human recruiters
\citep{jabarian-henkel-2026-voice-ai}. It is strong causal evidence for that
information-collection intervention in its setting, not evidence for automated
final selection or transportability to other labor markets and job families.

Other field studies show why deployment outcomes and human discretion must be
measured rather than assumed. Algorithmic worker recommendations increased
vacancy fill rates in one online labor market~\citep{horton-2017-algorithmic};
a retail job-test rollout linked screening to tenure and subgroup hiring
outcomes~\citep{autor-scarborough-2008-testing}; and higher apparent
manager-exception rates were associated with worse average hire outcomes in a
study spanning 15 service-sector firms~\citep{hoffman-etal-2018-discretion}.
These setting-specific results are stronger deployment evidence than offline
accuracy alone, but they concern
particular platforms, tests, jobs, and outcomes. They do not establish that an
LLM-based assessment is valid or that less human discretion is universally
preferable.

\subsection{Agent and profession-aligned benchmarks}

Agent benchmarks must log the environment, tools, model version, time, prompts,
actions, and evidence. xbench advances the field by collecting professional
recruiting tasks and connecting metrics to real workflow value
\citep{chen-etal-2025-xbench}. PeopleSearchBench advances factual verification
and multi-dimensional evaluation for returned people
\citep{shi-etal-2026-peoplesearchbench}. Together they mark a shift from
``Does the model classify this pair?'' to ``Can the system produce a usable,
verifiable work product?''

TalentTrace adds an upstream requirement lifecycle: 691 people requirements
are elicited, validated, searched, and verified across 21 systems
\citep{he-etal-2026-requirement-sourcing}. It exposes failures that begin before
retrieval, although union-pooled candidates and multi-model judging remain
system-dependent approximations to open-world recall.

The remaining challenges are substantial:

\begin{itemize}
  \item \textbf{Temporal reproducibility}: live pages and profiles change.
  Benchmarks need timestamps, archived evidence where lawful, and freshness
  metrics.
  \item \textbf{Open-world incompleteness}: the set of correct people is not
  enumerable. Scores need explicit pooling and coverage assumptions.
  \item \textbf{Judge dependence}: professional rubrics improve an LLM judge,
  but human validation and judge-replacement tests remain necessary.
  \item \textbf{Action safety}: research must distinguish proposing an outreach
  message from sending one, or finding public data from storing it.
  \item \textbf{Value versus validity}: labor saved on sourcing is an economic
  outcome, not proof of lawful or predictive employee selection.
\end{itemize}

\subsection{A tool trajectory is not a hiring lifecycle}

An agent tool trajectory covers search, tool calls, evidence collection,
shortlist production, and handoff. A hiring lifecycle continues through
recruiter contact, interview, offer, acceptance, start, performance, and
retention. The distinction matters because a locally safe trajectory can still
be evaluated with future information or against an outcome observed only for a
selected subset.

At checkpoint $k$, a defensible evaluation must restrict the system to
information $I_k$ that existed at the prediction timestamp. It should report
the eligible risk set, follow-up horizon, label-availability date, and handling
of censoring when a job closes, a candidate withdraws, or the outcome is not yet
observed. Market interference also matters: candidates compete for jobs, jobs
compete for candidates, recruiter policies change, and one system action can
alter later exposure. These features make lifecycle evaluation longitudinal,
partially observed, and often causal; they cannot be recovered by replaying a
short tool trace with information collected after the decision.

An end-to-end audit of roughly 497,000 candidate--vacancy pipeline records from
a public employment agency illustrates the value of this wider unit: aggregate
binary-gender parity coexisted with age, salary-level, and intersectional
disparities, while vendor opacity limited attribution of their cause
\citep{galdon-clavell-2026-applied-filtered}. This is longitudinal deployment
evidence, but missing qualification and downstream employer data prevent a
complete causal or selection-validity interpretation.

\subsection{Evidence requirements by claim level}

Evidence requirements follow the six evaluation units in
Figure~\ref{fig:evaluation-ladder}. L1--L6 are shorthand for those units, not a
second taxonomy, and should be chosen by the highest-consequence claim made:

\begin{enumerate}[label=\textbf{L\arabic*.}]
  \item \textbf{Field/document claim}: extraction correctness, provenance,
  critical-error severity, and source localization.
  \item \textbf{Pair claim}: a defined candidate--job label or criterion can be
  predicted, with label semantics, calibration, abstention, and subgroup error
  stated explicitly.
  \item \textbf{List claim}: full-corpus recall, shortlist ordering, stability,
  independently judged relevance, candidate-universe provenance, and slice
  coverage.
  \item \textbf{Case/selection claim}: criterion grounding, expert
  adjudication, job-related validity, selective deferral, differential
  prediction, adverse-impact analysis, and deployment-specific review.
  \item \textbf{Trajectory/workflow claim}: logged actions, verified
  completion, evidence freshness, latency, cost, recovery, permission tests,
  and human correction.
  \item \textbf{Outcome claim}: lifecycle analysis of contact, offer,
  acceptance, start, job-relevant performance, and retention with explicit
  risk sets, prediction timestamps, follow-up, delayed labels, censoring, and
  uncertainty. A causal claim additionally requires an assignment mechanism or
  stated identification assumptions, a causal estimand, and sensitivity or
  robustness analysis appropriate to the design.
\end{enumerate}

A paper need not claim every level. It must stop each conclusion at the most
consequential unit for which it supplies the corresponding evidence. The
numbering orders system scope and consequence; it does not imply that every
study at one unit is methodologically stronger than every study at another.
Fairness and security slices apply at each relevant unit rather than as one
aggregate audit appended at the end.

\section{Fairness, Privacy, Security, and Accountability}
\label{sec:governance}

Recruitment allocates access to livelihoods. Governance is therefore not a
post-processing module; it constrains what data may be collected, which
decisions may be automated, and what evidence is required before deployment.

\subsection{Fairness is stage- and task-dependent}

Counterfactual audits replace names or demographic signals while holding
qualifications constant. Studies report name-associated differences in hiring
and salary recommendations~\citep{an-etal-2024-large,
nghiem-etal-2024-gotta}, disability-related sensitivity in resume screening
\citep{glazko-etal-2024-disability}, and ranking changes after small demographic
perturbations~\citep{seshadri-etal-2025-small}. JobFair distinguishes mean-level
and dispersion-related bias and shows that the direction of group differences
can vary across settings~\citep{wang-etal-2024-jobfair}.

Related audits expand the comparison space. An industry-track study varies
gender, race, and education signals in job--resume matching
~\citep{iso-etal-2025-evaluating}; FAIRE evaluates direct scoring and ranking
under racial and gender perturbations~\citep{wen-etal-2025-faire}; and controlled
experiments identify model self-preference when the same model family both
writes and screens application text~\citep{xu-etal-2025-self-preference}.

These findings are not reducible to one model fairness score. Results depend on
the task (retrieval, scoring, generation, or ranking), population, prompt,
attribute proxy, model version, and metric. Removing a sensitive field is not a
complete solution because schools, employers, locations, language, and career
gaps can act as proxies. Conversely, parity on a synthetic name test does not
establish criterion validity.

Newer evidence reinforces the need for a testable audit design rather than a
single verdict. A 30,384-evaluation study found near-zero gender and race effects
under its clean paired-resume conditions while retaining positive controls,
equivalence tests, and prompt stress tests; small effects appeared under some
model and context combinations~\citep{pavlopoulos-2026-minimal}. Conversely, an
end-to-end public-employment audit found that aggregate binary-gender parity
could mask age, salary-level, and intersectional disparities across the
pipeline~\citep{galdon-clavell-2026-applied-filtered}. A risk-science review of
22 common fairness metrics further argues that severity estimates should be
reported with uncertainty about both occurrence and consequence
\citep{wilson-etal-2026-risk-science}. These studies ask different questions;
none is a portable fairness certificate.

Fairness-aware re-ranking in LinkedIn Talent Search demonstrates a different
intervention point: post-process a top-$K$ list toward a specified
representation distribution while preserving utility
~\citep{geyik-etal-2019-fairness}. The method is operationally concrete, but the
target distribution and protected-attribute policy remain normative choices.

A stage-aware audit asks where disparity enters:

\begin{itemize}
  \item \textbf{Exposure}: were comparable jobs or candidates shown?
  \item \textbf{Representation}: did parsing or normalization lose different
  evidence across document styles or languages?
  \item \textbf{Recall}: were qualified members retrieved at comparable rates?
  \item \textbf{Ranking}: did position or confidence differ after controlling
  for job-relevant evidence?
  \item \textbf{Human handoff}: did the system defer, and did humans override,
  at comparable rates?
  \item \textbf{Outcome}: did the deployed policy create adverse impact, and is
  the selection procedure demonstrably job-related?
\end{itemize}

Intersectional slices and uncertainty intervals are necessary because aggregate
parity can hide concentrated harms. Audit data should also reflect reasonable
accommodation: an assessment that is predictive on average may screen out a
person with a disability for reasons unrelated to ability to perform the job.
A disability-centered review additionally identifies a disclosure paradox:
adaptive or multimodal tools may improve access while voice, gaze, affect, and
outlier assumptions can penalize the same candidates or pressure them to reveal
sensitive information~\citep{guler-zhu-2026-disability}.

\subsection{Privacy, consent, and purpose limitation}

Resumes contain direct identifiers, education, employment history, inferred
seniority, contact information, and sometimes health or immigration-related
facts. Agents that search the open web add cross-context aggregation: facts that are public in
separate places can become intrusive when merged into a recruiting dossier.

An accountable system specifies the lawful and operational purpose for each
source, minimizes retention, separates search from storage, records source and
timestamp, and gives candidates a way to correct consequential inaccuracies.
De-identification does not itself authorize model training or eliminate proxy
attributes. Long-term agent memory requires particularly explicit boundaries:
information collected to satisfy one search should not silently become a
permanent candidate profile. These principles are operational duties in some
jurisdictions. China's Personal Information Protection Law addresses
transparency, fairness, explanation, and refusal for specified automated
decisions~\citep{china-2021-pipl}, while the General Data Protection Regulation
(GDPR) Article 22 provides safeguards for specified solely automated decisions,
including human intervention and the right to contest the decision
~\citep{eu-2016-gdpr}.

\subsection{Untrusted inputs, identity, and tool security}

Resumes, portfolio pages, search results, recruiter notes, and tool responses
cross a trust boundary when they enter an agent. The security problem is
broader than prompt injection:

\begin{itemize}
  \item \textbf{Instruction and ranking attacks}: embedded or hidden text can
  override criteria, inflate rank, reveal prompts, or trigger an action.
  Pre-LLM embedding rankers were already manipulable
  \citep{samadi-etal-2021-attacks}, and resume-borne instructions can change LLM
  screening outcomes~\citep{baxi-etal-2026-prompt}.
  \item \textbf{Identity and provenance failures}: entity resolution can merge
  two people, a stale page can overwrite a current record, and poisoned public
  evidence can enter a candidate dossier.
  \item \textbf{Cross-tool and memory failures}: retrieved personal data can
  leak into an unrelated tool call, an unauthorized outreach or ATS mutation
  can be executed, or malicious content can persist in long-term agent memory.
\end{itemize}

RAPIDS treats detection as a cascade and shows that a high-recall lightweight
detector followed by LLM verification can reduce cost and latency while
retaining attack coverage~\citep{augey-etal-2026-rapids}. Detection is only one
control. A defensible architecture separates data from instructions, binds
facts to source and timestamp, validates entity identity and schemas, grants
least-privilege tools, allowlists consequential actions, isolates case memory,
and requires authorization before outreach or record mutation. Evaluation
should report end-to-end attack success, detector recall and false-positive
burden, identity-collision error, stale-source handling, cross-tool leakage,
unauthorized-action prevention, and recovery after a rejected or partially
executed action.

LongPIBench extends this threat to long contexts and includes both synthetic and
real-world resume-screening cases. Its evaluations show that defenses appearing
effective on short inputs can be bypassed more often as document context grows
\citep{liu-etal-2026-longpibench}. This is benchmark evidence about defense
coverage, not a measured attack rate in production recruitment.

\subsection{Regulation and employment validity}

Technical evaluation should be aligned with, but not confused with, legal
compliance. Section~\ref{sec:industrial-regional} compares primary evidence from
China, the European Union, the United Kingdom, and the United States. The EU AI
Act lists specified recruitment and selection systems in Annex~III and
classifies covered uses through Article~6(2), subject to Article~6(3)
~\citep{eu-2024-ai-act}. The Article~6(3) derogation is unavailable when an
Annex~III system performs profiling of natural persons; such a system remains
high-risk. Regulation (EU) 2026/1744 was published in the
Official Journal on 24 July 2026 and entered into force on 27 July 2026; it sets
2 December 2027 as the application date for Chapter~III, Sections~1--3, except
Article~6(5), for systems classified as high-risk under Article~6(2) and
Annex~III~\citep{eu-2026-ai-omnibus}; Chinese rules directly
govern online recruitment services and specified algorithmic or automated-
decision practices~\citep{mohrss-2020-network-recruitment,
china-2022-algorithm-rules}; New York City requires covered automated
employment decision tools to undergo a bias audit and imposes publication and
notice duties~\citep{nyc-2021-local-law-144,nyc-dcwp-2023-aedt}; UK automated-
decision rules were amended by the Data (Use and Access) Act 2025
~\citep{uk-2025-duaa,uk-2026-duaa-commencement}; and the EEOC Uniform Guidelines
Q\&A distinguishes desirable validation from the Guidelines' validity-evidence
requirement triggered by adverse impact
~\citep{eeoc-1979-uniform-guidelines-qa}, while agency guidance is non-binding
and does not itself create legal obligations
~\citep{eeoc-2020-guidance-nonbinding}.

Two EU updates affect implementation before the delayed high-risk dates. AI Act
Article~50 has applied since 2 August 2026 and requires specified directly
interactive AI systems to inform people that they are interacting with AI
\citep{ec-2026-article50-guidelines}. Regulation (EU) 2026/1744 also added a
strictly conditioned Article~4a route for processing special-category data when
necessary to detect or correct bias; it requires safeguards and does not create
a general permission to rank on protected attributes or an independent duty to
run such processing~\citep{eu-2026-ai-omnibus}. Recent legal analysis likewise
centers effective remedies and access to justice for applicants rather than
treating a technical explanation alone as redress
\citep{parviainen-2026-remedies}.

These selected regimes differ, and this review does not offer legal advice or a
global legal census. Table~\ref{tab:regional-rules} is a time-stamped technical
map verified through \legalcutoff{}, not a durable compliance checklist. The
shared technical implication is that accuracy on an internal dataset is
insufficient. Obligations are actor-specific rather than interchangeable. For
a covered high-risk system once the relevant provisions apply, providers bear
system-compliance, quality-management, technical-documentation,
conformity-assessment, registration, corrective-action, and post-market
responsibilities. Deployers must use the system according to its instructions,
assign competent and empowered human oversight, monitor operation, retain logs
under their control, and satisfy applicable worker and affected-person
information duties~\citep{eu-2024-ai-act}. Other regimes allocate duties
differently, so contracts should not collapse vendor and employer
responsibilities into a single generic checklist. Earlier audits of
vendor claims similarly show why development, validation, and bias-mitigation
practices must be examined as a socio-technical process
\citep{raghavan-etal-2020-mitigating}.

\subsection{Accountability at system boundaries}

The phrase ``human in the loop'' does not specify responsibility. For every
stage, a deployment record should identify who owns the requirement, who can
change a score or filter, who approves an external action, how a candidate can
contest data, and what happens when sources disagree. Vendors, employers, and
recruiters cannot allocate these obligations to an abstract model.

Interviews with 21 hiring-tool developers and vendors illustrate why this
allocation matters: participants often described diversity and debiasing as a
later retrofit rather than a design input~\citep{yarrow-2026-inequality}.
The qualitative, predominantly US sample does not estimate industry prevalence,
but it makes developers and vendors visible as governance actors.

Model cards alone are also too narrow for compound systems. A useful recruitment
system card includes the retrieval corpus, source licenses, filter logic,
model and prompt versions, calibration set, known slice failures, human review
policy, tool permissions, retention schedule, and incident procedure.

\section{From Failure Modes to Intervention Paths}
\label{sec:solutions}

Within the reviewed evidence, risk diagnosis is stronger than proof of
end-to-end remedies. Still, several solution families are concrete enough to
organize. Table~\ref{tab:solution-paths} distinguishes three evidence states:
\emph{recruitment evidence} means the intervention has been tested in a
recruitment system or audit; \emph{transfer evidence} means the method is
established elsewhere but its recruitment effect is unproven; and
\emph{design synthesis} means the implementation proposal is derived from
multiple sources and remains to be validated.

\begingroup
\footnotesize
\setlength{\tabcolsep}{3.4pt}
\renewcommand{\arraystretch}{1.14}
\begin{longtable}{@{}P{2.75cm}P{4.65cm}P{3.8cm}P{4.5cm}@{}}
\caption{Problem--solution map with explicit evidence strength.}
\label{tab:solution-paths}\\
\toprule
\textbf{Failure mode} & \textbf{Existing or proposed path} &
\textbf{Evidence state} & \textbf{What must still be tested} \\
\midrule
\endfirsthead
\multicolumn{4}{l}{\textit{Table~\ref{tab:solution-paths} continued}}\\
\toprule
\textbf{Failure mode} & \textbf{Existing or proposed path} &
\textbf{Evidence state} & \textbf{What must still be tested} \\
\midrule
\endhead
\bottomrule
\endlastfoot

Exposure-biased clicks and applications & Estimate examination or exposure
propensities; use counterfactual or propensity-weighted learning and evaluation;
retain a small randomized logging policy where defensible
~\citep{schnabel-etal-2016-treatments,joachims-etal-2017-unbiased}. Model job
exposure and candidate exposure separately in a bilateral market. &
Transfer evidence from search and recommendation. &
Propensity misspecification, extreme weights, feedback loops, consent, and
whether a two-sided estimator improves qualified interviews rather than clicks. \\

Qualified targets absent from the pool & Combine lexical, dense, behavioral,
and graph recall; judge a target set independently; report recall before
re-ranking and the maximal reachable downstream score. &
Recruitment evidence for hybrid or dense retrieval, but limited open
full-corpus targets~\citep{yu-etal-2024-confit,yu-etal-2026-confit-v3}. &
Native-language coverage, rare credentials, cold starts, source availability,
and whether additional reach is equitable and useful. \\

Representation disparity in top-$K$ & Apply a declared target distribution in
fairness-aware re-ranking while constraining utility
~\citep{geyik-etal-2019-fairness}; audit exposure, recall, rank, and outcome as
separate stages. &
Recruitment production evidence for re-ranking. &
Who legitimately chooses the target, treatment of unknown attributes,
intersectional uncertainty, and downstream opportunity rather than list parity. \\

Proxy discrimination and opaque filters & Combine controlled counterfactual
tests, observed adverse-impact analysis, criterion validity, data-minimization
review, local-language job-ad wording scans, and periodic monitoring
~\citep{an-etal-2024-large,glazko-etal-2024-disability,
ico-2024-recruitment-audit,jiang-etal-2023-chinese-gender-lexicon,
pavlopoulos-2026-minimal,galdon-clavell-2026-applied-filtered}. Remove or
redesign a filter only after documenting its purpose and assessing less
discriminatory alternatives. &
Recruitment audit evidence; remediation-effect evidence is sparse. &
Whether mitigation transfers from synthetic pairs to the deployed population,
preserves job relevance, and remains stable under model and market drift. \\

Unsupported scores or rationales & Represent requirements as atomic criteria;
link every consequential claim to a source span; distinguish supported,
contradicted, and unknown; abstain when evidence is insufficient
~\citep{buchmann-etal-2024-attribute,castleman-etal-2026-validity}. &
Mixed recruitment and long-document evidence. &
Independent expert correctness, omission detection, user correction, and the
effect of evidence-first interfaces on recruiter error rather than preference. \\

Unsafe or decorative human review & Learn or define a rejector that routes
cases to a competent reviewer when the expected utility of human review exceeds
that of automation
~\citep{mozannar-sontag-2020-defer,verma-nalisnick-2022-calibrated}. Combine it
with mandatory routing for sensitive, conflicting, or externally acting cases. &
Transfer evidence plus early recruitment-specific intervention evidence
~\citep{takano-2026-fabrication,zhou-etal-2026-rejected}. &
Reviewer heterogeneity, calibration, automation bias, workload, differential
deferral, override quality, and accountability for the final action. \\

Untrusted inputs, identity failures, and unsafe actions & Cascade a high-recall
detector with stronger verification~\citep{augey-etal-2026-rapids}; stress-test
long contexts~\citep{liu-etal-2026-longpibench}; bind facts
to source, time, and resolved identity; isolate document data and case memory
from instructions; use typed tool calls, least privilege, action allowlists,
and confirmation before outreach or record mutation. &
Recruitment-specific attack and detector evidence; identity, memory, and
full-workflow defenses remain design synthesis. &
Adaptive and indirect attacks, false-positive burden, entity collisions, stale
or poisoned evidence, cross-tool leakage, memory contamination, recovery, and
end-to-end attack success under realistic permissions. \\

\end{longtable}
\endgroup

Fairness Hazard Analysis offers a complementary requirements-engineering path:
map process nodes, actors, propagated hazards, and candidate controls before a
model is deployed~\citep{broccia-etal-2026-fairness-hazard}. Its focus-group and
organizational case evidence supports feasibility and practitioner usefulness,
not a claim that the proposed mitigations improve hiring outcomes.

\subsection{A testable human-handoff contract}

A compliant handoff should be a protocol rather than a button labeled
``human review.'' We propose the following design synthesis for evaluation:

\begin{enumerate}
  \item \textbf{Trigger}: route on calibrated uncertainty, missing mandatory
  evidence, source conflict, protected or sensitive information, policy
  exceptions, and any irreversible external action.
  \item \textbf{Evidence bundle}: show the request, retrieved pool, applied
  filters, criterion-level sources, uncertainty, model and prompt version, and
  the exact action proposed.
  \item \textbf{Authority}: identify a reviewer competent for the decision and
  give that reviewer the ability to request information, override, or stop the
  workflow; passive approval is not meaningful oversight.
  \item \textbf{Execution boundary}: separate recommendation from execution.
  Outreach, interview or assessment dispatch, rejection, offer generation,
  profile persistence, and applicant tracking system (ATS) updates require
  explicit authority and are logged as distinct actions.
  \item \textbf{Contest and learning}: record the reason for override, expose a
  correction or challenge channel where required, and evaluate the joint
  human--AI policy rather than retraining blindly on every human decision.
\end{enumerate}

For implementation and audit, the handoff should be serialized as a versioned
record containing the case and request identifiers, candidate entity
resolution, pool and filter provenance, criterion-level claims and sources,
uncertainty, proposed action, required authority, reviewer decision and reason,
timestamps, and recovery state. The schema makes the contract testable without
assuming that a rendered approval screen is faithful to the underlying trace.

This contract is intentionally presented as an outlook, not a validated
recruitment architecture. A credible study would compare it with ordinary
review under a fixed human-time budget and measure decision quality, correction
quality, latency, automation bias, subgroup deferral, and candidate appeals.
The DfE vacancy-QA transparency record supplies one concrete pre-deployment
pattern---risk-tiered sampling, per-check false-negative priority, and
fail-to-human routing---but not yet a validated general recruitment architecture
\citep{uk-dfe-2026-vacancy-qa}.

\section{Research Agenda}
\label{sec:agenda}

The next advances are unlikely to come from replacing every component with a
larger model. The higher-value problems concern supervision, evaluation,
system boundaries, and causal impact.

\subsection{Separate preference, qualification, and opportunity}

Behavior logs collapse several latent variables. Future datasets should model
at least candidate preference, employer qualification, exposure, and process
opportunity as distinct signals. Reciprocal recommendation can then optimize a
constrained objective rather than one engagement score. Research should report
how conclusions change under alternative negative-sampling and exposure models.
Counterfactual recommendation and unbiased learning-to-rank provide estimators
for biased observation processes~\citep{schnabel-etal-2016-treatments,
joachims-etal-2017-unbiased}; the recruitment-specific challenge is to model
exposure and response on both sides without treating either propensity model as
known.

\subsection{Build open, multilingual, full-pipeline benchmarks}

The reviewed public benchmarks often start from clean text or a supplied
shortlist, while
real systems begin with heterogeneous documents and a large candidate universe.
A full-pipeline benchmark should include native PDFs, OCR or layout variants,
multilingual job descriptions, hard and preferred criteria, independently
judged recall targets, criterion-level evidence, partial orders among
candidates, and allowable actions. The design must separate synthetic control
from real-world external validity rather than asking one dataset to provide
both.

Chinese and other non-English settings are particularly important. Translation
of an English benchmark tests cross-lingual model transfer but does not recreate
local credentials, labor-market conventions, platform behavior, or regulatory
context. Multilingual work should therefore combine parallel cases with
culturally native cases and report the distinction.

\subsection{Evaluate counterfactual reachability, not only final accuracy}

For a target set $T$ and recalled pool $R_K$, the maximum attainable target
recall is
\begin{equation}
\operatorname{Recall}_{\max}(R_K;T)=\frac{|T\cap R_K|}{|T|}.
\end{equation}
Other downstream metrics require their own assumptions about ordering, utility,
and missing relevance judgments. Benchmarks should publish eligible reach,
maximal reachable performance, and error attribution by stage. Counterfactual
interventions can replace an OCR output, retrieval pool, or evidence set with
gold while holding downstream components fixed. This identifies the slow or
binding stage and prevents every failure from being attributed to the final
LLM.

\subsection{Move from plausible explanations to verified evidence}

Recruitment explanations should be decomposed into atomic claims, linked to
source spans or dated external evidence, and checked for completeness as well
as correctness. Unknown and contradictory evidence must be represented
explicitly. Research should compare whether evidence-first interfaces improve
human error detection, not merely whether users prefer longer rationales.

\subsection{Design selective and budgeted human collaboration}

The goal is not maximum automation at any cost. A system should learn which
cases to resolve, which fact to acquire, and which cases to route to a human
under a review budget. Evaluation should compare uncertainty-based deferral,
rule-based critical-risk routing, and learned policies on the same budget.
Human studies should measure independent judgment, automation bias, correction
quality, time, and differential deferral across groups.
Learning-to-defer methods supply a formal starting point
~\citep{mozannar-sontag-2020-defer,verma-nalisnick-2022-calibrated}, but a
recruitment policy must also route mandatory legal or policy exceptions that
cannot be reduced to model uncertainty.

\subsection{Treat temporal validity as a first-class property}

Candidate roles, publications, locations, and availability change. Open-source
sourcing benchmarks need a time-indexed evidence protocol: query timestamp,
source timestamp, archived or hashed evidence where permitted, freshness
threshold, and explicit handling of conflict. Agent leaderboards should preserve
both a stable anchor set and a rotating live set so that longitudinal model
progress is not confounded completely by task replacement.

\subsection{Measure causal utility and long-term effects}

Offline ranking improvements do not answer whether the system increases
qualified interviews, reduces time without reducing quality, improves candidate
experience, or changes workforce composition. When deployment is possible,
staged experiments should track multiple funnel outcomes, correction and appeal,
and longer-term criteria such as retention or performance---with clear warnings
about post-treatment selection. Where randomization is infeasible, researchers
should state the identification assumptions of quasi-experimental designs.
Every checkpoint should freeze the information available at that time, specify
the eligible population and follow-up window, and report delayed labels and
censoring. Competing candidates, jobs, recruiter policies, and market shocks
create interference; a complete lifecycle study therefore needs a longitudinal
estimand, not only a successful agent tool trace.

\subsection{Evaluate multi-objective trade-offs jointly}

In the coded set, quality, utility, cost, fairness, and security appear in
different combinations; privacy is not directly evaluated, and no row directly
evaluates all six axes. These are set-specific observations, not field-frequency
estimates. Compound-system design requires Pareto analysis. A small high-recall
detector, dense retriever, strong re-ranker, and selectively invoked LLM may
dominate an all-LLM design on cost and safety. Evaluations should report
quality--latency--cost curves alongside
slice performance and attack robustness, with failure budgets for high-severity
events.

\subsection{Make benchmarks contestable}

Benchmarks themselves encode a theory of good recruiting. Domain experts should
co-design tasks, but candidates, labor researchers, disability experts, and
legal or ethics specialists should also be able to challenge requirements,
labels, and metrics. Versioned benchmark cards should record whose values are
represented, what consequential use is excluded, and how errors can be
reported. Professional alignment without stakeholder contestability risks
optimizing the productivity of one actor at the expense of another.

Table~\ref{tab:agenda} consolidates these questions and the minimum evidence
needed to support their corresponding claims.

\begin{table}[H]
\centering
\small
\caption{Priority research questions and the minimum evidence needed.}
\label{tab:agenda}
\begin{tabularx}{\textwidth}{@{}P{3.5cm}Y@{}}
\toprule
\textbf{Research question} & \textbf{Minimum credible evidence} \\
\midrule
Does an LLM improve matching? & Fixed full-corpus recall, strong ranker
baselines, independent graded judgments, and slice analysis \\
Does an agent improve recruiting work? & Logged trajectories, verified evidence,
task completion, cost/latency, and human correction \\
Can the system make selection decisions? & Job-related criterion validity,
calibration, abstention, adverse-impact analysis, and deployment-specific review \\
Does human review make it safe? & Randomized interface study, blind baseline,
automation-bias measures, and authority to override \\
Does a live benchmark track progress? & Versioned tasks, timestamps, stable
anchors, rotating live cases, and confidence intervals \\
Does a deployed system improve employment outcomes? & Eligible risk set;
offer, start, job-relevant performance, and retention; prediction timestamps,
follow-up, censoring, assignment or identification assumptions, and uncertainty \\
\bottomrule
\end{tabularx}
\end{table}

\section{Limitations}
\label{sec:limitations}

This review has five scope and evidence limitations. First, the 40 coded works
form a purposive map of contrasting system objects, signals, evaluation units,
and risk surfaces. They are not a statistically representative sample, and the
review does not estimate the prevalence of methods or findings. Second, the
item-level coding is an author interpretation and was not independently
dual-coded; the companion file exposes those judgments for inspection rather
than supplying inter-rater reliability evidence.

Third, the primary technical discovery surfaces are predominantly
English-language, while Chinese-language and China-situated evidence was added
through a targeted supplement. This improves regional specificity but does not
constitute broad multilingual or global coverage. Fourth, much of the
2025--2026 evidence for recruiting agents comes from preprints, benchmarks, and
system demonstrations. It establishes emerging artifacts and evaluation
problems more strongly than stable deployment effects, and reported results may
change with model, tool, judge, or Web versions.

Fifth, private data and confidential experiments limit independent validation
of production claims and obscure negative results, subgroup errors, costs, and
long-term employment outcomes. The regional legal comparison is likewise a
technical map checked through \legalcutoff, not legal advice or a durable
compliance checklist. These limitations motivate claim-specific boundaries:
they restrict generalization without invalidating the source-specific evidence
summarized in the review.

\section{Conclusion}

Recruitment AI has evolved from rule-based search and bilateral recommendation
to behavioral ranking, neural semantic matching, LLM-augmented pipelines, and
tool-using agents. The progression is not a clean replacement sequence. Mature
systems still need lexical recall, structured constraints, inexpensive rankers,
and human approval; LLMs add flexible interpretation and action but also new
sources of uncertainty and control failure.

The methodological consequence is a staged evidence chain: an upstream result
can bound what follows, but cannot validate a later decision unit by itself.
xbench and open people-search benchmarks are valuable because they move
evaluation toward professional work products and factual utility. Employment
claims still require construct validity, temporal controls, stage-level
diagnostics, human collaboration, outcome evidence, and governance audits.
Recent randomized interview evidence and an end-to-end public-employment audit
show that such higher-level studies are feasible, while also demonstrating how
strongly their interpretation depends on workflow, population, human authority,
and missing downstream labels~\citep{jabarian-henkel-2026-voice-ai,
galdon-clavell-2026-applied-filtered}.

The public evidence base remains incomplete, but it need not be
one-dimensional. Industry papers expose architectures and online objectives;
engineering disclosures and annual reports expose product scope and geographic
commitment; regulator audits expose operational defects; and regional rules
constrain data, automation, oversight, notice, and contestability. Keeping these
sources separate while triangulating them is more informative than either
discarding non-academic evidence or accepting corporate claims at face value.
The solution literature supports concrete intervention families---
counterfactual learning, hybrid recall, fairness-aware re-ranking, grounded
abstention, selective deferral, and privilege-separated action---but not a claim
that the full recruitment problem is solved.

These conclusions remain bounded by the review design in
Section~\ref{sec:limitations}: the selected map is not a field-prevalence
estimate, and the time-stamped legal comparison is not a compliance checklist.

The resulting standard is demanding but claim-specific: a credible recruiting
agent should find relevant candidates without erasing the long tail,
distinguish unknown from negative evidence, ground consequential claims, expose
the limits of its search, preserve candidate preference and correction rights,
defer selectively, and separate recommendation from authorized action. Progress
is not the production of a more persuasive hiring recommendation. It is the
production of a more reliable, reciprocal, contestable, and useful recruitment
process.

\section{Materials Availability}
\label{sec:materials}

The companion files \texttt{literature\_map\_coding.csv} and
\texttt{supplementary\_search\_log.csv} are distributed with the manuscript
source bundle. The present manuscript release is version 1.13, dated 3 September
2026. The files' SHA-256 checksums are, respectively,
\nolinkurl{423a4a8ed2431d105b3dbe5623d9f60958c5262a970b684aa52664b06a3f4f84}
and
\nolinkurl{376f4235d04f801d72b2e1f14694e17193b09f7a4ec7092700b54a57512c9828}.
These checksums identify the reviewed source artifacts in this release; field
definitions and coding boundaries are documented in
Section~\ref{sec:protocol}.

\appendix
\section{Recruitment-AI Evidence Landscape}
\label{app:evidence-landscape}

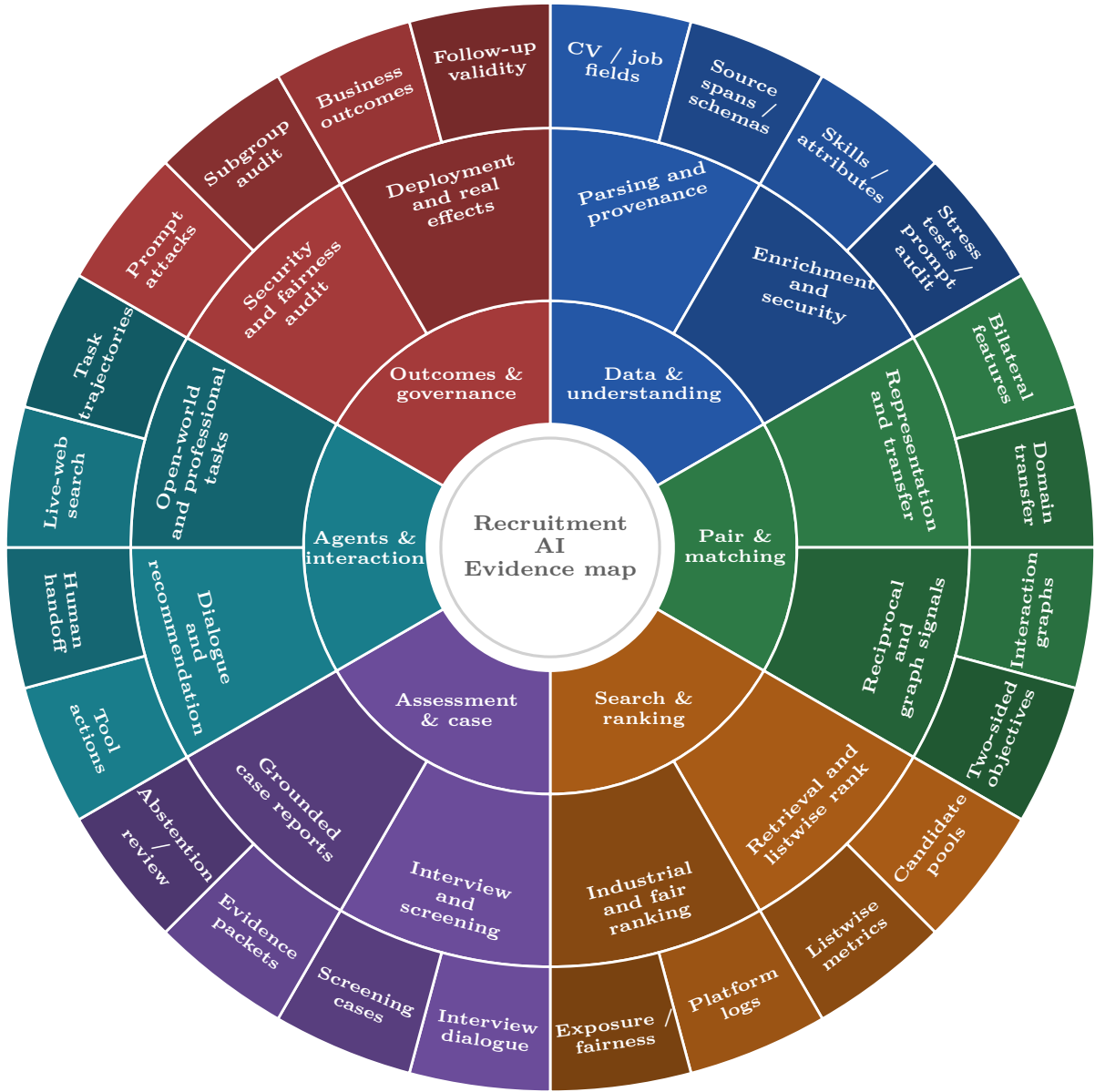
\begin{figure}[H]
\centering
\begingroup
\hyphenpenalty=10000\relax
\exhyphenpenalty=10000\relax
\newcommand{\midouterradius}{5.10}
\newcommand{\middlelabelradius}{4.45}
\newcommand{\outerlabelradius}{5.95}
\newcommand{\annular}[5]{%
  \path[fill=#5, draw=white, line width=0.9pt]
    (#1:#3) arc[start angle=#1,end angle=#2,radius=#3] --
    (#2:#4) arc[start angle=#2,end angle=#1,radius=#4] -- cycle;
}
\resizebox{0.96\textwidth}{!}{%
\begin{tikzpicture}[x=1cm,y=1cm,
  innerlabel/.style={text=white, font=\fontsize{6.2}{6.8}\selectfont\bfseries,
    align=center, inner sep=0pt},
  middlelabel/.style={text=white, font=\fontsize{6.2}{6.8}\selectfont\bfseries,
    align=center, text width=2.20cm, inner sep=0pt},
  outerlabel/.style={text=white, font=\fontsize{6.0}{6.6}\selectfont\bfseries,
    align=center, text width=1.50cm, inner sep=0pt}]

\annular{90}{30}{1.50}{3.00}{stageblue}
\annular{30}{-30}{1.50}{3.00}{stagegreen}
\annular{-30}{-90}{1.50}{3.00}{stageorange}
\annular{-90}{-150}{1.50}{3.00}{stagepurple}
\annular{-150}{-210}{1.50}{3.00}{stagecyan}
\annular{-210}{-270}{1.50}{3.00}{stagered}

\annular{90}{60}{3.00}{\midouterradius}{stageblue}
\annular{60}{30}{3.00}{\midouterradius}{stageblue!80!black}
\annular{30}{0}{3.00}{\midouterradius}{stagegreen}
\annular{0}{-30}{3.00}{\midouterradius}{stagegreen!80!black}
\annular{-30}{-60}{3.00}{\midouterradius}{stageorange}
\annular{-60}{-90}{3.00}{\midouterradius}{stageorange!80!black}
\annular{-90}{-120}{3.00}{\midouterradius}{stagepurple}
\annular{-120}{-150}{3.00}{\midouterradius}{stagepurple!80!black}
\annular{-150}{-180}{3.00}{\midouterradius}{stagecyan}
\annular{-180}{-210}{3.00}{\midouterradius}{stagecyan!80!black}
\annular{-210}{-240}{3.00}{\midouterradius}{stagered}
\annular{-240}{-270}{3.00}{\midouterradius}{stagered!80!black}

\annular{90}{75}{\midouterradius}{6.62}{stageblue}
\annular{75}{60}{\midouterradius}{6.62}{stageblue!82!black}
\annular{60}{45}{\midouterradius}{6.62}{stageblue!92!black}
\annular{45}{30}{\midouterradius}{6.62}{stageblue!72!black}
\annular{30}{15}{\midouterradius}{6.62}{stagegreen}
\annular{15}{0}{\midouterradius}{6.62}{stagegreen!82!black}
\annular{0}{-15}{\midouterradius}{6.62}{stagegreen!92!black}
\annular{-15}{-30}{\midouterradius}{6.62}{stagegreen!72!black}
\annular{-30}{-45}{\midouterradius}{6.62}{stageorange}
\annular{-45}{-60}{\midouterradius}{6.62}{stageorange!82!black}
\annular{-60}{-75}{\midouterradius}{6.62}{stageorange!92!black}
\annular{-75}{-90}{\midouterradius}{6.62}{stageorange!72!black}
\annular{-90}{-105}{\midouterradius}{6.62}{stagepurple}
\annular{-105}{-120}{\midouterradius}{6.62}{stagepurple!82!black}
\annular{-120}{-135}{\midouterradius}{6.62}{stagepurple!92!black}
\annular{-135}{-150}{\midouterradius}{6.62}{stagepurple!72!black}
\annular{-150}{-165}{\midouterradius}{6.62}{stagecyan}
\annular{-165}{-180}{\midouterradius}{6.62}{stagecyan!82!black}
\annular{-180}{-195}{\midouterradius}{6.62}{stagecyan!92!black}
\annular{-195}{-210}{\midouterradius}{6.62}{stagecyan!72!black}
\annular{-210}{-225}{\midouterradius}{6.62}{stagered}
\annular{-225}{-240}{\midouterradius}{6.62}{stagered!82!black}
\annular{-240}{-255}{\midouterradius}{6.62}{stagered!92!black}
\annular{-255}{-270}{\midouterradius}{6.62}{stagered!72!black}

\draw[fill=white, draw=black!18, line width=1pt] (0,0) circle (1.33cm);
\node[align=center, text width=2.28cm,
  font=\fontsize{7.4}{8.0}\selectfont\bfseries, text=black!62] at (0,0)
  {Recruitment AI\\Evidence map};

\node[innerlabel] at (60:2.28) {Data \&\\understanding};
\node[innerlabel, text width=1.35cm] at (0:2.25)
  {Pair \&\\matching};
\node[innerlabel] at (-60:2.28) {Search \&\\ranking};
\node[innerlabel] at (-120:2.28) {Assessment\\\& case};
\node[innerlabel, text width=1.52cm] at (180:2.25)
  {Agents \&\\interaction};
\node[innerlabel] at (-240:2.28) {Outcomes \&\\governance};

\node[middlelabel, rotate=15] at (75:\middlelabelradius)
  {Parsing and\\provenance};
\node[middlelabel, rotate=-15] at (45:\middlelabelradius)
  {Enrichment\\and\\security};
\node[middlelabel, rotate=-75] at (15:\middlelabelradius)
  {Representation\\and transfer};
\node[middlelabel, rotate=75] at (-15:\middlelabelradius)
  {Reciprocal\\and\\graph signals};
\node[middlelabel, rotate=45] at (-45:\middlelabelradius)
  {Retrieval and\\listwise rank};
\node[middlelabel, rotate=15] at (-75:\middlelabelradius)
  {Industrial\\and fair\\ranking};
\node[middlelabel, rotate=-15] at (-105:\middlelabelradius)
  {Interview\\and\\screening};
\node[middlelabel, rotate=-45] at (-135:\middlelabelradius)
  {Grounded\\case reports};
\node[middlelabel, rotate=-75] at (-165:\middlelabelradius)
  {Dialogue\\and\\recommendation};
\node[middlelabel, rotate=75] at (-195:\middlelabelradius)
  {Open-world\\and professional\\tasks};
\node[middlelabel, rotate=45] at (-225:\middlelabelradius)
  {Security\\and fairness\\audit};
\node[middlelabel, rotate=15] at (-255:\middlelabelradius)
  {Deployment\\and real\\effects};

\node[outerlabel, rotate=-7.5] at (82.5:\outerlabelradius) {CV / job\\fields};
\node[outerlabel, rotate=-22.5] at (67.5:\outerlabelradius)
  {Source\\spans /\\schemas};
\node[outerlabel, rotate=-37.5] at (52.5:\outerlabelradius)
  {Skills /\\attributes};
\node[outerlabel, rotate=-52.5] at (37.5:\outerlabelradius)
  {Stress tests /\\prompt\\audit};
\node[outerlabel, rotate=-67.5] at (22.5:\outerlabelradius) {Bilateral\\features};
\node[outerlabel, rotate=-82.5] at (7.5:\outerlabelradius) {Domain\\transfer};
\node[outerlabel, rotate=82.5] at (-7.5:\outerlabelradius) {Interaction\\graphs};
\node[outerlabel, rotate=67.5] at (-22.5:\outerlabelradius) {Two-sided\\objectives};
\node[outerlabel, rotate=52.5] at (-37.5:\outerlabelradius) {Candidate\\pools};
\node[outerlabel, rotate=37.5] at (-52.5:\outerlabelradius) {Listwise\\metrics};
\node[outerlabel, rotate=22.5] at (-67.5:\outerlabelradius) {Platform\\logs};
\node[outerlabel, rotate=7.5] at (-82.5:\outerlabelradius) {Exposure /\\fairness};
\node[outerlabel, rotate=-7.5] at (-97.5:\outerlabelradius) {Interview\\dialogue};
\node[outerlabel, rotate=-22.5] at (-112.5:\outerlabelradius) {Screening\\cases};
\node[outerlabel, rotate=-37.5] at (-127.5:\outerlabelradius) {Evidence\\packets};
\node[outerlabel, rotate=-52.5] at (-142.5:\outerlabelradius) {Abstention /\\review};
\node[outerlabel, rotate=-67.5] at (-157.5:\outerlabelradius) {Tool\\actions};
\node[outerlabel, rotate=-82.5] at (-172.5:\outerlabelradius) {Human\\handoff};
\node[outerlabel, rotate=82.5] at (-187.5:\outerlabelradius) {Live-web\\search};
\node[outerlabel, rotate=67.5] at (-202.5:\outerlabelradius) {Task\\trajectories};
\node[outerlabel, rotate=52.5] at (-217.5:\outerlabelradius) {Prompt\\attacks};
\node[outerlabel, rotate=37.5] at (-232.5:\outerlabelradius) {Subgroup\\audit};
\node[outerlabel, rotate=22.5] at (-247.5:\outerlabelradius) {Business\\outcomes};
\node[outerlabel, rotate=7.5] at (-262.5:\outerlabelradius) {Follow-up\\validity};

\end{tikzpicture}%
}
\endgroup
\caption{Recruitment-AI evidence landscape. The inner ring groups six system
objects, the middle ring separates recurring mechanisms and evidence questions,
and the outer ring identifies the principal data, task, metric, and evidence
forms used to instantiate them. Sector size is an organizing device, not a
field-frequency estimate. Table~\ref{tab:literature-map} maps all 40 coded works
to system objects, reported evidence, and claim boundaries using author--year
citations.}
\label{fig:evidence-landscape}
\end{figure}

Figure~\ref{fig:evidence-landscape} provides a compact topology of the
representative literature map. System objects occupy the inner ring, recurring
mechanisms and evidence questions occupy the middle ring, and the outer ring
identifies the principal data, task, metric, and evidence forms. Sector sizes
are an organizing device and do not estimate field prevalence. The figure is
placed in the appendix because Table~\ref{tab:literature-map} carries the
source-specific evidence and claim boundaries used in the main synthesis.

\footnotesize
\setlength{\bibsep}{0pt}
\bibliographystyle{plainnat}
\bibliography{references}

\end{document}